\documentclass[10pt,twocolumn,letterpaper]{article}

\usepackage{cvpr}
\usepackage{times}
\usepackage{epsfig}
\usepackage{graphicx}
\usepackage{amsmath}
\usepackage{amssymb}

\usepackage{subfigure}
\usepackage{booktabs}
\usepackage{ntheorem}
\theoremseparator{:}
\newtheorem{hyp}{Hypothesis}
\usepackage{paralist}

\usepackage[pagebackref=true,breaklinks=true,letterpaper=true,colorlinks,bookmarks=false]{hyperref}

\cvprfinalcopy 

\def\cvprPaperID{10106} 

\ifcvprfinal\fi

\begin{document}

\title{The Impact of Hole Geometry on Relative Robustness of In-Painting Networks: An Empirical Study}

\author{Masood Mortazavi \\
Futurewei, Inc.\\
Santa Clara\\
{\tt\small masood.mortazavi@futurewei.com}
\and
Ning Yan\\
Futurewei, Inc.\\
Santa Clara\\
{\tt\small yan.ningyan@futurewei.com}
}

\maketitle

\begin{abstract}
In-painting networks use existing pixels to generate appropriate pixels to fill ``holes" placed on parts of an image. 
A 2-D in-painting network's input usually consists of 
\begin{inparaenum}[(1)]
\item a three-channel 2-D image, and 
\item an additional channel for the ``holes" to be in-painted in that image. 
\end{inparaenum}
In this paper, 
we study the robustness of a given in-painting neural network against variations in hole geometry distributions. 
We observe that the robustness of an in-painting network is dependent on the probability distribution function (PDF) of the hole geometry presented to it during its training even if the underlying image dataset used (in training and testing) does not alter.
We develop an experimental methodology for testing and evaluating relative robustness of in-painting networks against four different kinds of hole geometry PDFs.
We examine a number of hypothesis regarding \begin{inparaenum}[(1)]
\item the natural bias of in-painting networks to the hole distribution used for their training,
\item the underlying dataset's ability to differentiate relative robustness as hole distributions vary in a train-test (cross-comparison) grid, and
\item the impact of the directional distribution of edges in the holes and in the image dataset.
\end{inparaenum}
We present results for $L_1$, $PSNR$ and $SSIM$ quality metrics and develop a specific measure of relative in-painting robustness to be used in cross-comparison grids based on these quality metrics. (One can incorporate other quality metrics in this relative measure.)  
The empirical work reported here is an initial step in a broader and deeper investigation of ``filling the blank" neural networks' sensitivity, robustness and regularization with respect to hole ``geometry" PDFs, and it suggests further research in this domain.    
\end{abstract}


\newcommand{\rulesep}{\unskip\ \vrule\ }

\section{Introduction}
In everyday life, human beings have to cope with visual occlusions, auditory noise and other sensory gaps. 
Cognitive scientists have concluded that perception relies heavily on generative and predictive completion in active interaction with the environment\cite{clark2016surfing}. 
Image in-painting refers to using existing, ``valid" pixels (the pixel ``context") to generate appropriate pixels to fill ``holes," ``gaps" or ``blanks" in certain areas of an image. Usually, these ``holes" are placed where an image editor would like to remove certain unwanted parts of an image and in-paint these wholes appropriately given the context the ``valid" pixels provide.  
Image in-painting is also a sub-category in the more general category of ``filling the blank" tasks\cite{FedusGD18}.

One method of in-painting involves copying candidate patches where a few dominating patches could provide information for completing the missing parts\cite{He_statisticsof}.  
In-painting has been examined from the perspective of feature representations\cite{Pathak2016ContextEF}, along with joint tasks such as segmentation when in-painting of occluded objects \cite{ehsani2018segan}, or general in-painting in computer vision and photography\cite{liu2018partialinpainting, yu2018generative, yu2018free}. 
It is common to use an encoder-decoder DNN (including skip connections \textit{a la} UNET\cite{Ronneberger2015UNetCN}) in order to generate an in-painted image\cite{IizukaSIGGRAPH2017, liu2018partialinpainting}. The in-painting network architecture we used in our experiments can be said to belong to the UNET ``family" of network architectures---with some very important distinction which we discuss later in this paper. 
One recent proposal uses foreground-aware image in-painting system that first detects and completes the contours of the foreground objects in the image, then uses the completed contours as a guidance to in-paint the image\cite{Xiong_2019_CVPR}.
Another recent proposal uses region-wise convolutions to reconstruct existing and missing regions separately and then integrates this with a non-local operation to model a global correlation between existing and missing regions, leading to a coarse-to-fine framework for in-painting the image\cite{ijcai2019-433}.

Various architectural techniques, e.g., 
dilated convolutions\cite{IizukaSIGGRAPH2017}, 
partial convolutions\cite{liu2018partialinpainting, liu2018partialpadding}, 
gated convolutions\cite{yu2018free}, 
attention\cite{yu2018generative},  and
self-attention\cite{ImageTransfomer} have been deployed.
GAN-training schemes with adversarial losses\cite{IizukaSIGGRAPH2017,yu2018generative}, 
as well as perceptual, style and total variation losses\cite{liu2018partialinpainting} 
have all been used and compared\cite{yu2018free, yu2018generative}. 
It is also worth noting that theoretical and framework implementations for image transformation and synthesis\cite{wang2018pix2pixHD} can be easily extended to frameworks to train and test various in-painting networks.   

Some researchers have already pointed out that the hole geometry selection (in training and in test) can make a significant difference to the quality of in-painting but there are no specific published works exclusively focused on exploring the impact of the hole geometry on the relative robustness of in-painting networks. 
The hole to be in-painted is represented by a 2-D binary channel, fed along with the ``masked" image to the in-painting network\cite{Pathak2016ContextEF, liu2018partialinpainting}.   
Some investigators have used one or more rectangular masks\cite{Pathak2016ContextEF, yu2018generative, IizukaSIGGRAPH2017}. In some of these cases, the network architecture is not ``task-invariant". For example, the intentional mechanism in \cite{yu2018generative} allows only rectangular holes. 
Other investigations use a larger variety of hole geometries during training\cite{liu2018partialinpainting, yu2018free}. 
In \cite{liu2018partialinpainting}, in particular, occlusion/dis-occlusion estimation between two consecutive frames for videos described in \cite{DensePointOptFlow2010} have been used as a source for hole patterns. 
When training an in-painting network, these patterns are subjected to random cropping, rotation, re-sizing and other similar operations to produce a usable and largely random distribution of holes. 
Finally, there is a proposal that seeks to optimize ``holes" provided by the image editors by developing and using ``naturalness'' metrics\cite{Isogawa2018} based on the idea of ``super-pixel", the collection of locally and similarly colored pixels which respect object boundaries. (Super-pixels seem to have also been used as a standard working unit of pre-CNN-based segmentation algorithms\cite{Vezhnevets05}.) 

While most published work have come to a consensus that non-rectangular holes need to be used during training, no one has given a clear review or rigorous enough analysis or account of how hole geometries used during training can impact the relative robustness of in-painting networks. In this paper, we take some steps towards providing such an analysis and review. 
The present paper contains the following contributions: 
\begin{inparaenum}[(1)]
\item We devise an experimental evaluation method for cross-comparison of the robustness of a given network when trained on a variety of hole probability distribution functions (PDFs). (Specifically, in this study, we use four such distributions.)  
\item We examine a set of hypothesis regarding the impact of hole geometry on the robustness of a given in-painting network. 
\item We show that when holes are aligned with convolution kernels, less robust networks are trained.
\item We show that the underlying image datasets can differentiate relative in-painting hole-robustness to different degrees, and by way of explanation, we show some correlation between differentiation capacity of a dataset with its edge gradient distribution.
\item Finally, we describe and use a heuristic approach to combine various quality metrics in order to measure the impact of in-training hole geometry distribution on the ``relative robustness'' of in-painting networks when subjected to a set of hole distributions. 
\end{inparaenum}
We observe that our \textit{cross-comparison} methodology can be extended to evaluate the impact of ``blank" distribution on relative robustness of other input-output systems designed for ``filling the blank" in other sensory data, including text, audio and speech. 
%
%
%
%
%

\section{Background and Experimental Setup} \label{experimental}

\begin{figure}
    \centering
    \includegraphics[width=\linewidth]{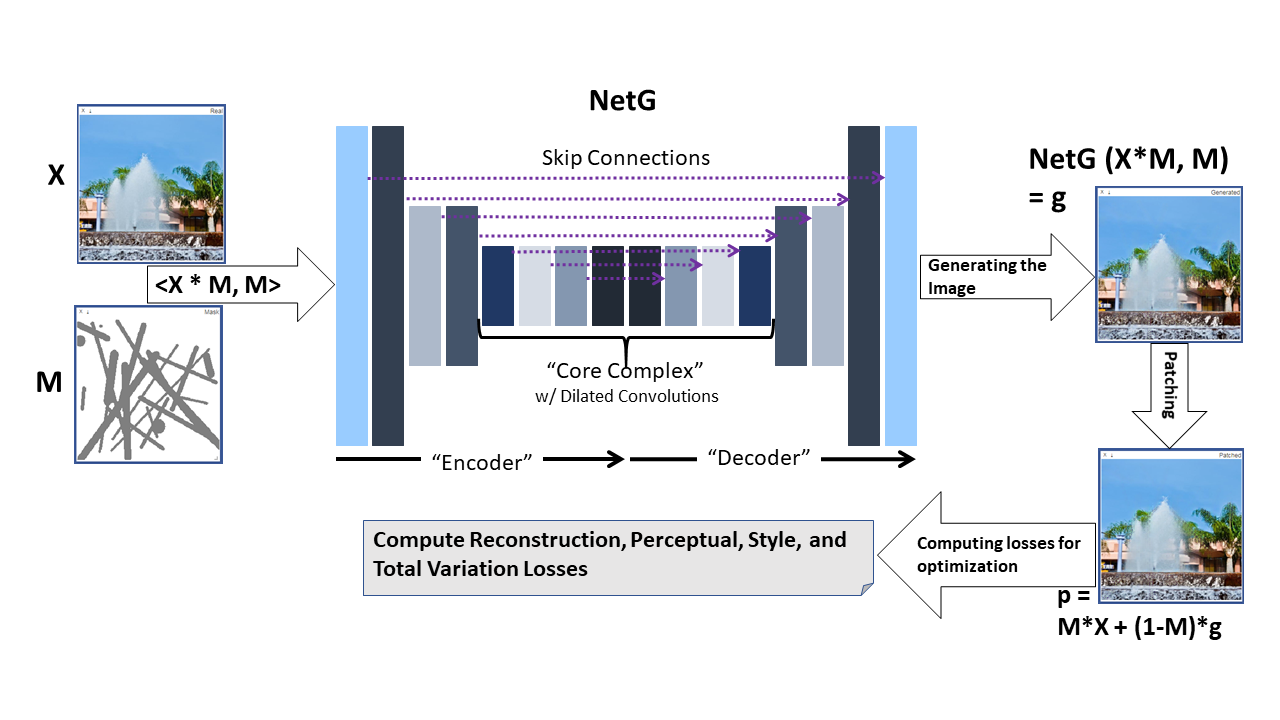}
    \caption{A typical neural in-painting architecture, in \textit{inference perspective}.} 
    \label{fig:architecture}
\end{figure}

Figure \ref{fig:architecture} shows our in-painting network architecture. For brevity and focus, we only name losses in summary form and do not show branches where they are computed.  In obtaining the results discussed in this paper, we limited the losses used (i.e., the optimization objectives) to those described in \cite{liu2018partialinpainting}.  
 
\subsection{A Typical In-Painting Architecture} \label{inference_arch}
In-painting networks' input requires the original image $X$ along with the hole ``mask'' $M$. 
(The network is fed the tuple $<X.M, M>$.) 
The hole mask, $M$, is a 2-D, single-channel tensor of binary values of the same array height ($H$) and width ($W$) as the image itself. 
The binary value at position $<x,y>$ of the $M$ is $0$ if the image needs to be ``in-painted'' in that position.  It is $1$ otherwise. 

The in-painting neural networks we designed are in the UNET\cite{Ronneberger2015UNetCN} family. 
The network used to produce the results in this paper has $16$ layers---eight layers in the encoder and eight layers in the decoder. 
There \textit{are} some major differences with the classical UNET: In our experimental network, there are only two convolution layers (layer 3 and layer 5) which involve contractions of feature maps in the encoder (as in \cite{liu2018partialinpainting,IizukaSIGGRAPH2017}). 
The number of convolution kernels (output feature maps) are organized according to a scaling factor $N_{fm}$, as $N_{fm}*(1, 2, 2, 4, 4, 4, 4, 4)$ in the encoder. The numbers of decoder kernels are symmetrically arranged, i.e. $N_{fm}*(4, 4, 4, 4, 4, 2, 2, 1)$.  
%
%
%
%
%
%
In the central $8$ layers of our networks ($4$ in the encoder and $4$ in the decoder) we have a \textit{core-complex} with a dilation pattern applied to convolutions. (This gives a far better receptive field in the core at the cost of higher working memory usage given current implementations of dilated convolution operators.) 
The networks we have used in this paper all use partial convolutions in their encoder phase. In the decoder, we use regular convolutions. 
It should be clear that the manner in which $M$ (the input hole pattern) contracts in the encoder (in Figure \ref{fig:architecture}) for any given convolution architecture will be fixed and known when using partial convolutions\cite{liu2018partialinpainting}. 
(Here, we note that in gated convolutions\cite{yu2018free}, this ``hole" contraction pattern, along the network, is trainable.) 
Neither the network architectures we examined and used nor the training and inference schemes we have employed impose any restrictions on hole geometry. 
As long as the hole can be expressed in a single-channel pixel format of $1$ (for no hole) and $0$ (for hole), the pattern can be used. 

We have developed in-painting networks in the range of sizes for various applications. 
For this paper, we focused on a small (10MB, $N_{fm} = 32$) in-painting neural network that generalizes well on common 256x256 image datasets. 

\subsection{Hole Geometry Distributions Used in Training} \label{holes}

Table \ref{tab:color_scheme} identifies the acronyms for the hole geometries we have used for reference throughout the paper. 
\begin{table}
\begin{center}
\scalebox{0.85}
{
\begin{tabular}{|l|l|}
\hline
Description of Hole Distribution & Acronym \\
\hline\hline
Random Rectangles & $RR$\\
\hline
Randomly Rotated $RR$ & $R4$\\
\hline
Random Geometric Shapes & $RGS$\\
\hline
Random Flow-based Occlusion/de-Occlusion Masks  & $RFOM$ \\
\hline
\end{tabular}
}
\end{center}
\caption{Acronyms used for each hole distribution.}
\label{tab:color_scheme}
\end{table}

\begin{figure*}[h]
    \centering
    \subfigure[RR]{
    \includegraphics[width=0.07\linewidth]{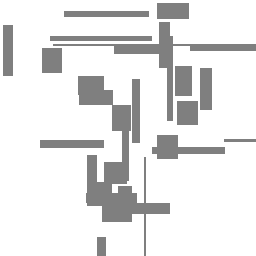}
    \rulesep
    \includegraphics[width=0.07\linewidth]{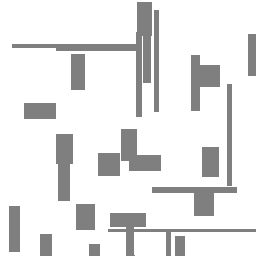}
    \rulesep
    \includegraphics[width=0.07\linewidth]{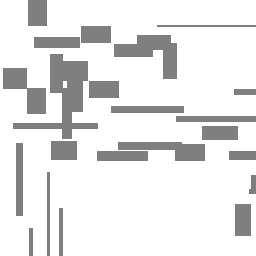}
    }
    \rulesep
    \subfigure[R4]{
    \includegraphics[width=0.07\linewidth]{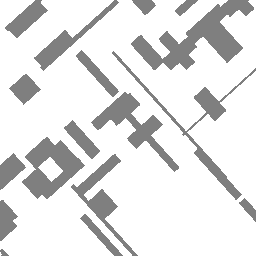}
    \rulesep
    \includegraphics[width=0.07\linewidth]{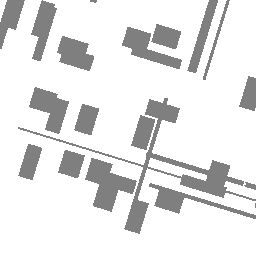}
    \rulesep
    \includegraphics[width=0.07\linewidth]{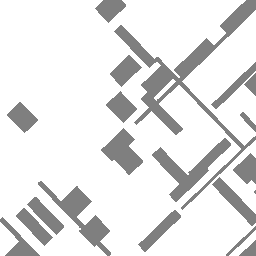}
    }
    \rulesep
    \subfigure[RGS]{
    \includegraphics[width=0.07\linewidth]{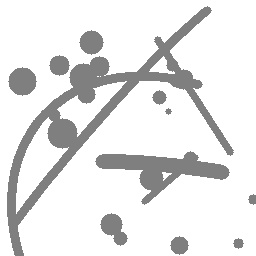}
    \rulesep
    \includegraphics[width=0.07\linewidth]{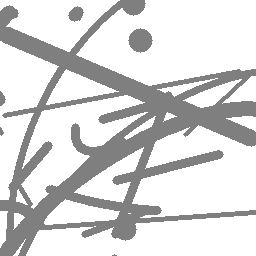}
    \rulesep
    \includegraphics[width=0.07\linewidth]{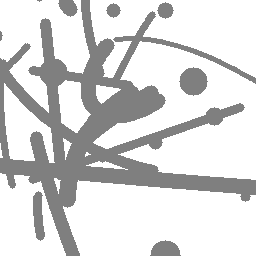}
    }
    \rulesep
    \subfigure[RFOM]{
    \includegraphics[width=0.07\linewidth]{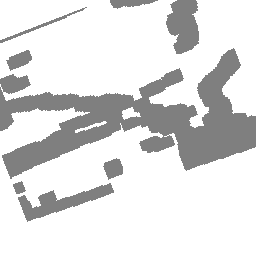}
    \rulesep
    \includegraphics[width=0.07\linewidth]{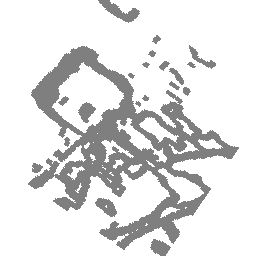}
    \rulesep
    \includegraphics[width=0.07\linewidth]{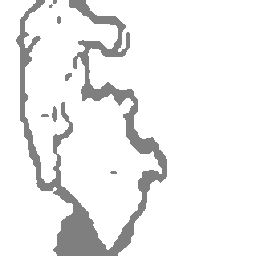}
    }
    \centering
    \caption{Three sample masks from each of the four hole geometry distributions examined.}
    \label{fig:mask_samples}
\end{figure*}

Our four distributions include random rectangles ($RR$), randomly rotated random rectangles ($R4$), random geometric shapes ($RGS$), and random flow-based occlusion/dis-occlusion masks ($RFOM$). 
Figure \ref{fig:mask_samples} gives some typical samples of hole distributions that were used in training our in-painting networks. 

Note that $RFOM$ uses flow-based occlusion/de-occlusion observed in some consecutive video frames\cite{DensePointOptFlow2010, liu2018partialinpainting}. 
Sample feed files for these flow-based hole masks have been released\cite{irregular_masks, liu2018partialinpainting}. 
These require pre-processing by randomly applying rotations, flips, re-size and crops. In all cases, we set our hole generation parameters to ensure similar distribution of the hole sizes. Random rotations might cause some variations of roughly the same order.   

\subsection{Datasets and Training Schedules} \label{training}

To keep training and testing used in this report relatively short, manageable and easily repeatable by others, we used the validation set in MIT places\cite{NIPS2014Places}. 
Our experiments with the full MIT Places dataset take much longer. Although some of $RR$-trained networks bias differential for $RR$ disappear, it can still be observed, and the conclusions and hypothesis in section \ref{results} still hold. (See the supplementary material.) 
As a second dataset, we also used a 90/10 (training/test) split of Celeb-A\cite{liu2015faceattributes}. 
For testing, we used a random selection of 1024 images in the corresponding test datasets. 
Using larger set of test images from the test datasets made little difference in the relative position or shape of the quality metric histograms we have reported in section \ref{results}. In fact, we observed the same trends (as section \ref{results}) when using much smaller tests sets---as small as 128 images. 

For deep-learning training and test platform, we used PyTorch. Our hardware consisted of NVIDIA 1080-ti and V100 GPUs.     
We trained our networks for two epochs with constant learning rate and for another epoch with uniformly diminishing learning rate, using the Adam optimizer, and the losses and hyper-parameters as set in \cite{liu2018partialinpainting}. 
Mini-batches consisted of 32 images and 32 holes (randomly selected or generated for each image, according to one of the four ``hole" distributions).
The trends, reported in section \ref{results}, remain largely the same even if we trained networks for a larger number of epochs, or trained larger networks of similar structure but larger number of kernels, i.e., larger $N_{fm}$. (See supplementary results.)

\subsection{Testing} \label{testing}


In this study, we are interested in relative robustness of multiple versions of a given network. 
A network trained with any of the hole geometry distributions is tested against all hole geometry distributions, including the one used for its own training. 
We propose \textit{cross-comparison}, i.e., cross-examination in a train-test grid as a method for measuring \textit{relative robustness} with respect to hole geometry. 

\subsection{Quality of In-painting} \label{q_metrics}

Strict quantitative analysis of the quality of in-painting can be difficult \cite{yu2018generative}.  
Nevertheless, it is still possible to formulate a quantitative method for comparative analysis of the \textit{relative robustness} of in-painting networks (although strict quality judgment would still require examination by human eyes). 

For purposes of such quantitative measures of relative robustness, we used $L_1$ (or reconstruction error\cite{Pathak2016ContextEF}), $PSNR$ (Peak Signal-to-Noise Ratio), $SSIM$ (Structural SIMilarity\cite{Wang04imagequality}). We did not use $IS$ (Inception Score\cite{ImprovedGAN}) given that there have been some recent critiques of $IS$ as measure of quality\cite{inception2018note}. However, the method we have developed here can be easily extended to incorporate $IS$, or other measures of quality. 

\subsection{''Holistic'' relative robustness} \label{sec:holistic}

Motivated by our findings (see section \ref{results}), we formulated a (``holistic'') measure for relative robustness of in-painting networks trained with respect to varying hole distributions. 

Relative robustness expressions $\mathcal{R}_M$ can be formulated for any given quality metric, $M$. 
For $PSNR$ and $SSIM$, whose higher value generally means better reconstruction, we define  in-painting ``relative robustness" (based on these individual quality metrics) as below:
\begin{equation}\label{eq:m}
\mathcal{R}_M(net_r) = (\sum_{i \ne r}^{}\bar{M}(i, net_r))
\end{equation}
where $M$ stands for either the $PSNR$ or $SSIM$ metrics, $r$ is a specific hole distribution $i$. We use $net_r$ (and sometimes $r_{train}$) to indicate that a network has been trained using hole distribution with index (or acronym) $r$. 
The index $i$ iterates over all such hole distributions, and $\bar{M}(i, net_r)$ is the expected value of $M$  when $net_r$ is tested against hole distribution $i$. 
Here, a bar over a value expresses the statistical expectation of that value.

In the case of $L_1$, since lower values mean better reconstruction, the relative robustness,$\mathcal{R}_{L_1}$ , can be defined as follows: 
\begin{equation}\label{eq:L_1}
\mathcal{R}_{L_1}(net_r) = \sum_{i \ne r}^{}1/\bar{L_1}(i, net_r)
\end{equation}
Here, $\bar{L_1}(i, net_r)$ is the expected $L_1$ loss when $net_r$ is tested against hole  distribution $i$.

We experimented with various ways to combine test metrics and found that the sums just described, above, provide the best indicators. 

A ``holistic" (and heuristic) measure of relative robustness can then be obtained by combining the above relative robustness measures based on quality metrics:
\begin{equation}\label{eq:holistic}
\mathcal{R}_{holistic}(net_r) = \sum_{m}\alpha(m)*\mathcal{R}_m(net_r)
\end{equation}
where $m$ is any of the metrics.
The appropriate setting of $\alpha(m)$ weights can balance the contribution of each metric $m$ in the ``holistic'' measure. 
We do not currently have a recipe for setting these metric-specific weights. 
Nevertheless, in the values reported here, we use the following scheme in order to give them equal weight in computing $\mathcal{R}_{holistic}(net_r)$:
\begin{equation}\label{eq:alpha}
\alpha(m) = 1 / \max\limits_{net_i} \mathcal{R}_m(net_i)
\end{equation}
This choice of weights, $\alpha(m)$, normalizes the contribution (to ``relative robustness") of each metric with respect to the best of each quality metric $\mathcal{R}_m$ to $1$, and all the other values to some value between $0$ and $1$. (See the right most column in Table \ref{tab:means_table}.)

We close this section with some final remarks: 
\begin{inparaenum}[(1)]
\item While ``holistic'' measures of relative robustness such as the one proposed here may be indicative, other measures can also be used or combined in the same format for cross-comparison purposes. 
\item Reviewing, more closely, the theory of network robustness in the case of unsupervised/semi-supervised training (for unlabled datasets) may also help us propose better measures of robustness.    
\end{inparaenum}
We have left tackling these items for future work.

\section{Results} \label{results}
In this section, we evaluate the relative robustness of an in-painting network with respect to hole geometry distribution. 

In our results, we refer to a network trained using a hole distribution $r$ as $r_{train}$, or $net_r$, where $r$ is the index or acronym for the hole distribution. 

We used $net_r$ in the equations discussed in section \ref{sec:holistic} but in the text in this section, for better legibility of hole type used for training, we will use $r_{train}$. For example, $RGS_{train}$ (and $net_{RGS}$) refers to a network trained with the $RGS$ hole distribution. 

\subsection{Cross-comparisons based on PDFs of quality metrics} \label{his}

\begin{figure*}[h]
    \centering
    \subfigure[Test w/ RandomRectangles - $RR$ hole configuration. $L_1$, $PSNR$, $SSIM$ histograms. Left: Places. Right: Celeb-A] 
    {
        \includegraphics[width=0.5\linewidth]{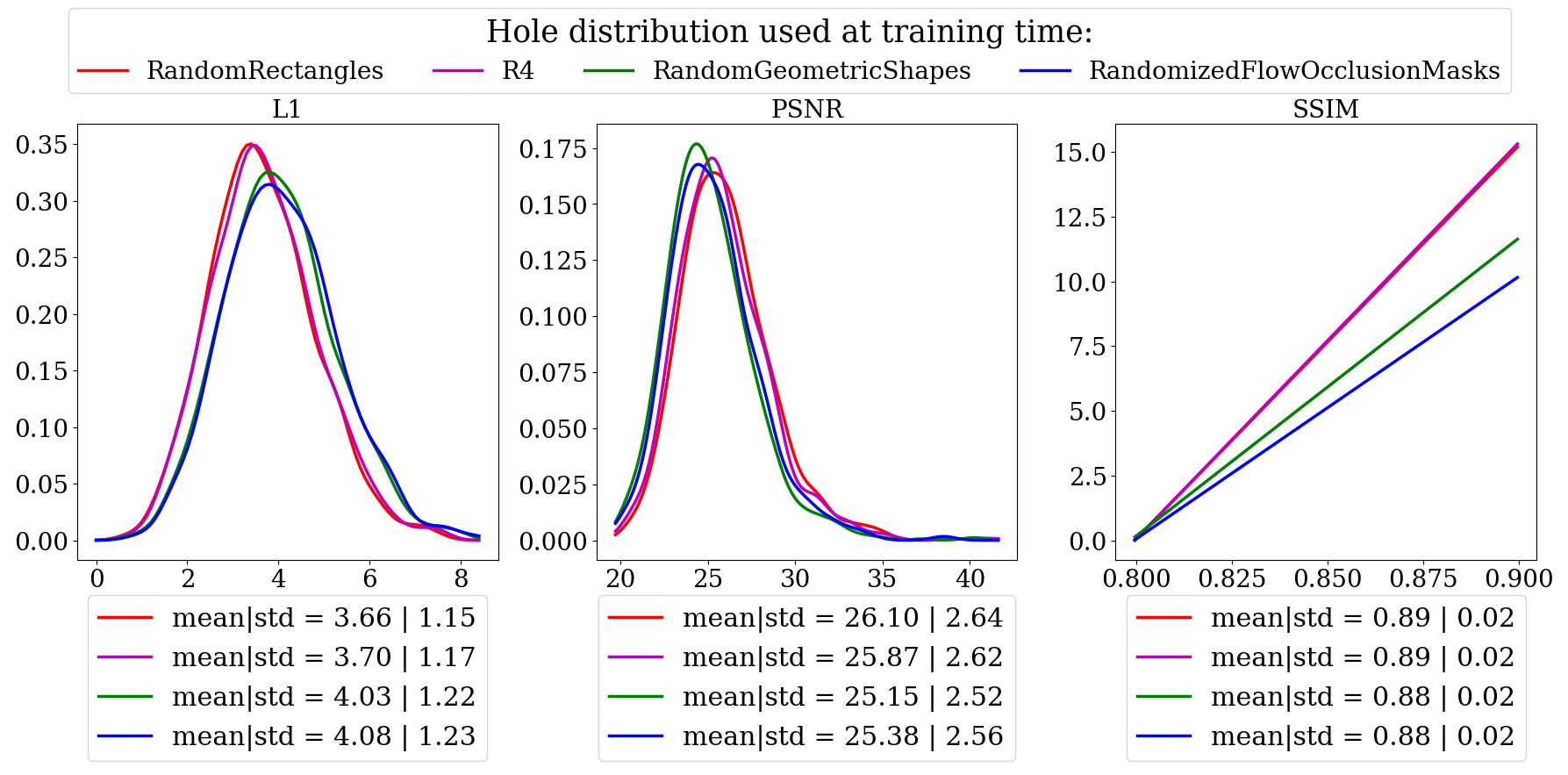}
    	\rulesep
        \includegraphics[width=0.5\linewidth]{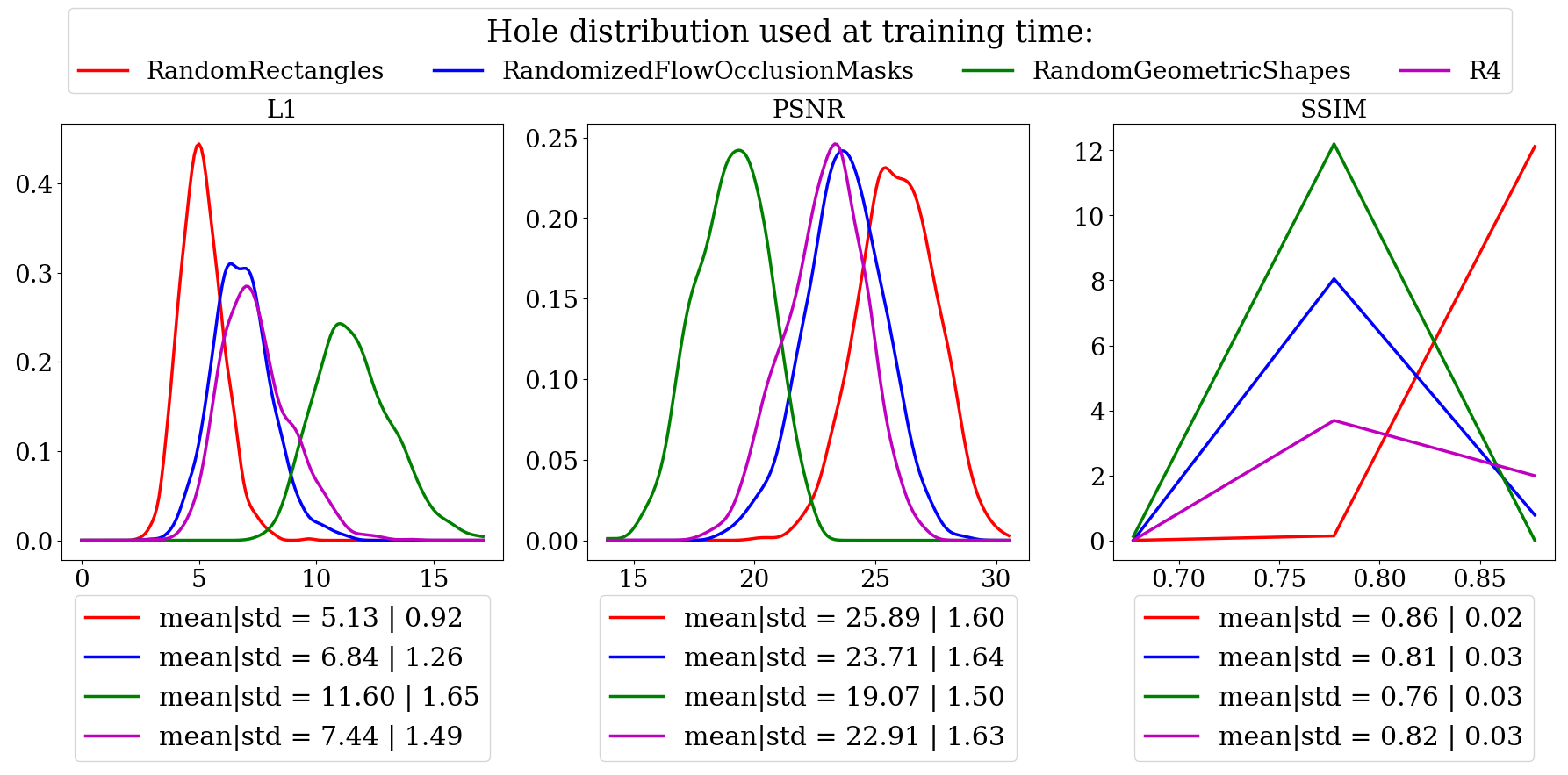}
        \label{pf_f32_rr}
    }
    \\
    \subfigure[Test w/ Randomly Rotated $RR$ - $R4$ hole configuration. $L_1$, $PSNR$, $SSIM$ histograms. Left: Places. Right: Celeb-A] 
    {
        \includegraphics[width=0.5\linewidth]{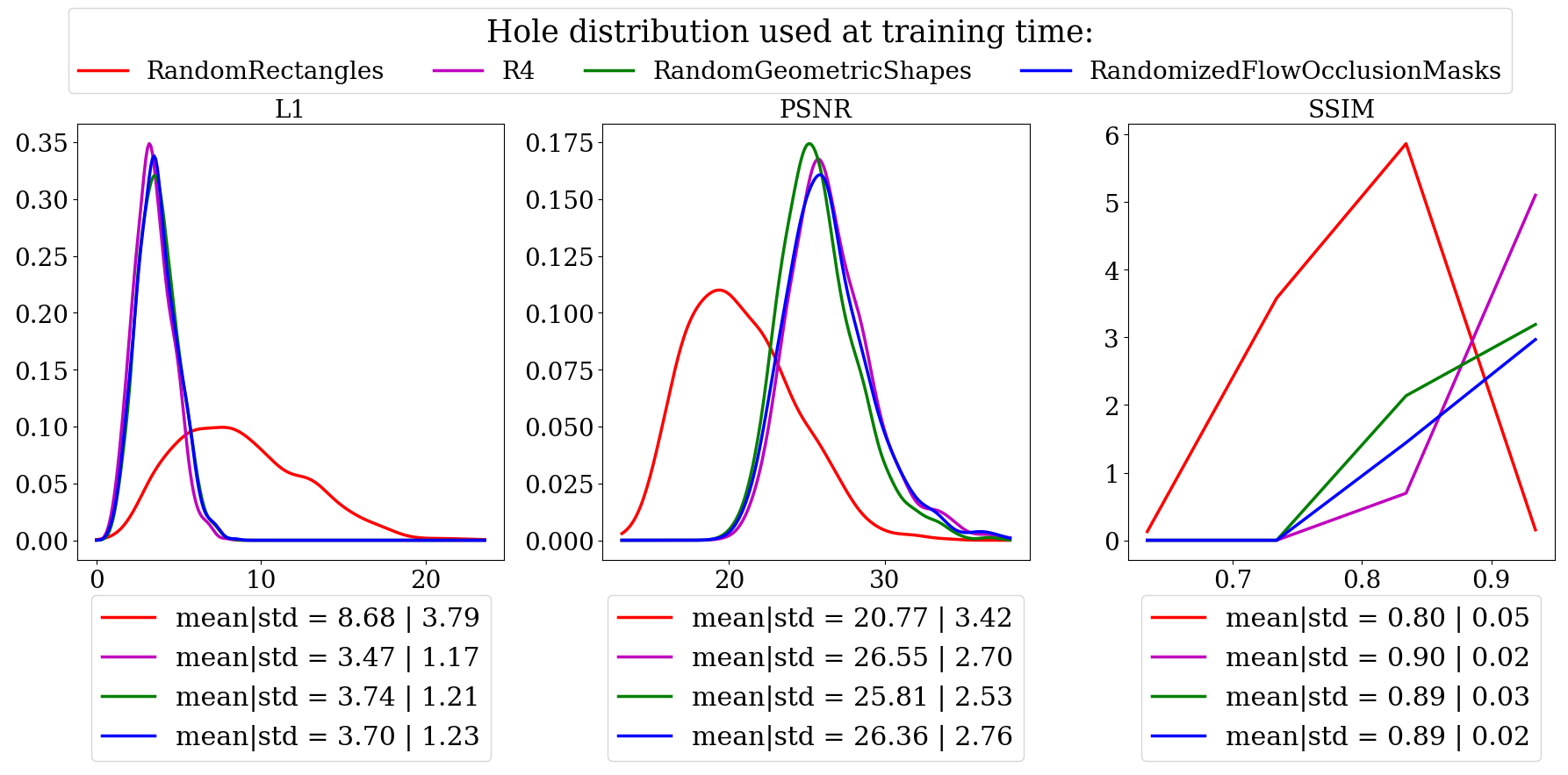}
    	\rulesep
        \includegraphics[width=0.5\linewidth]{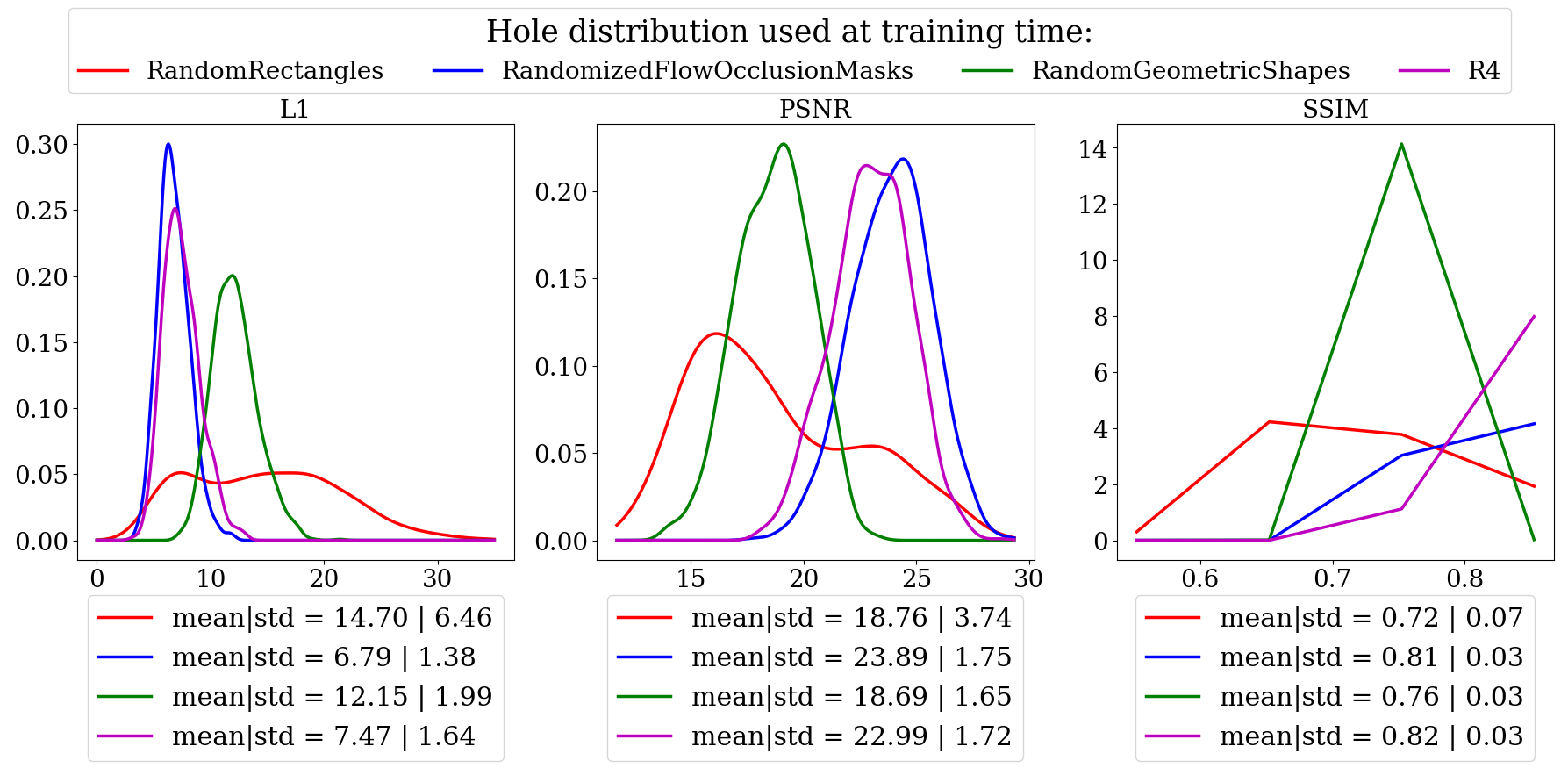}
        \label{pf_f32_r4}
    }
    \\
    \subfigure[Test w/ Random Geometric Shapes - $RGS$ hole configuration. $L_1$, $PSNR$, $SSIM$ histograms. Left: Places. Right: Celeb-A] 
    {
        \includegraphics[width=0.5\linewidth]{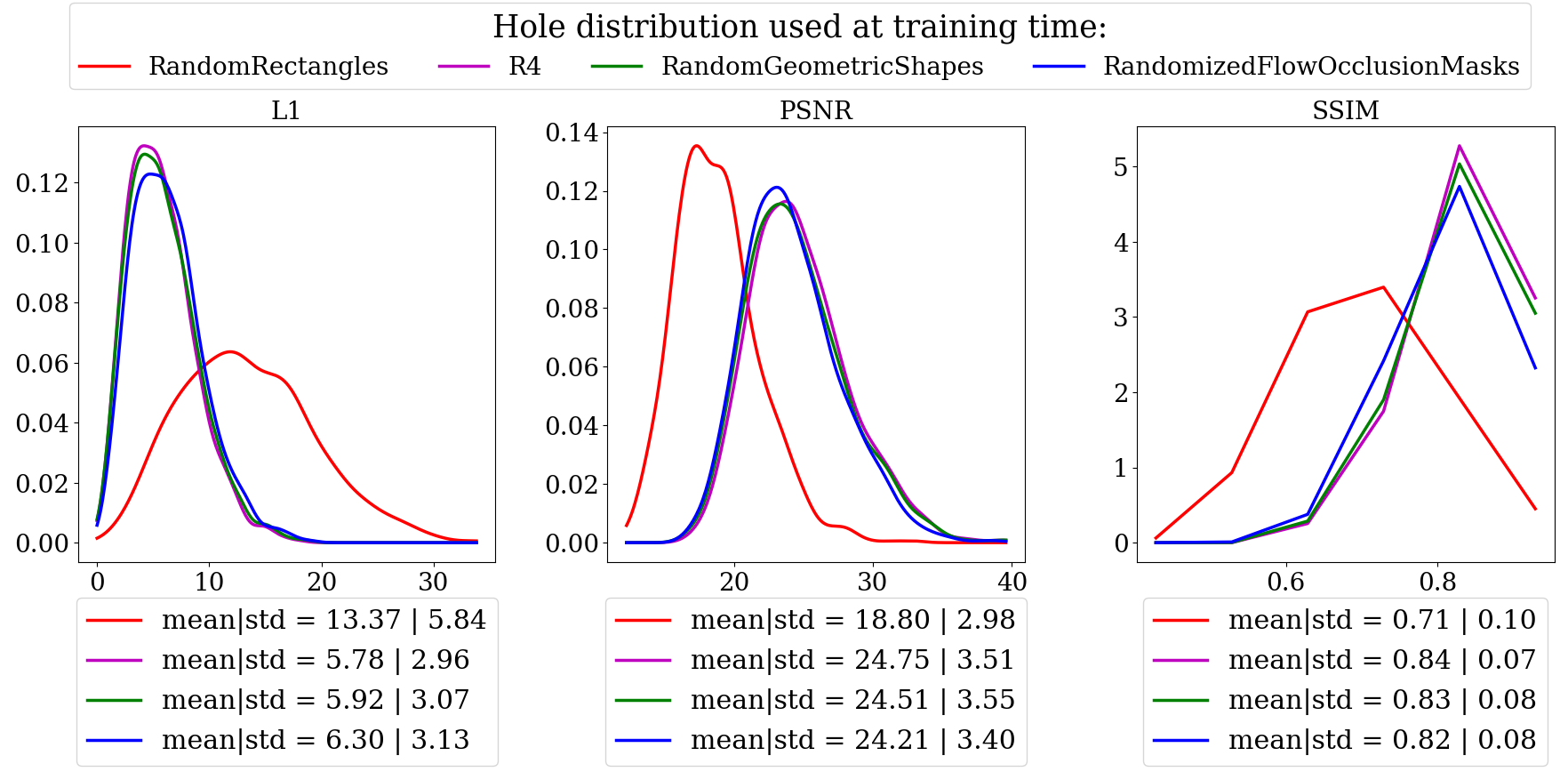}
    	\rulesep
        \includegraphics[width=0.5\linewidth]{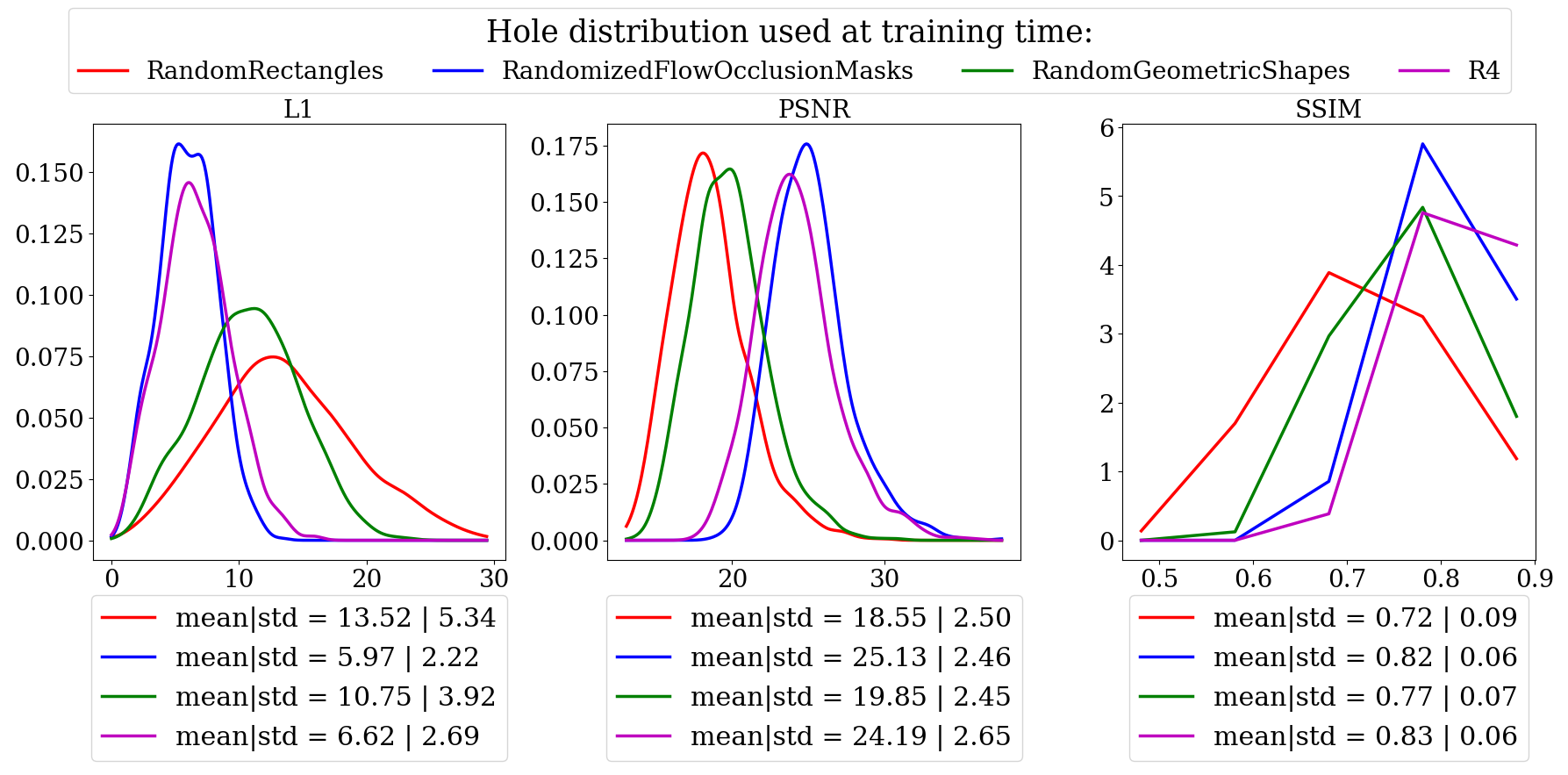}
        \label{pf_f32_rgs}
    }
    \\
    \subfigure[Test w/ Random Flow-based Occlusion/de-Occlusion Masks - $RFOM$ hole configuration. $L_1$, $PSNR$, $SSIM$ histograms. Left: Places. Right: Celeb-A] 
    {
        \includegraphics[width=0.5\linewidth]{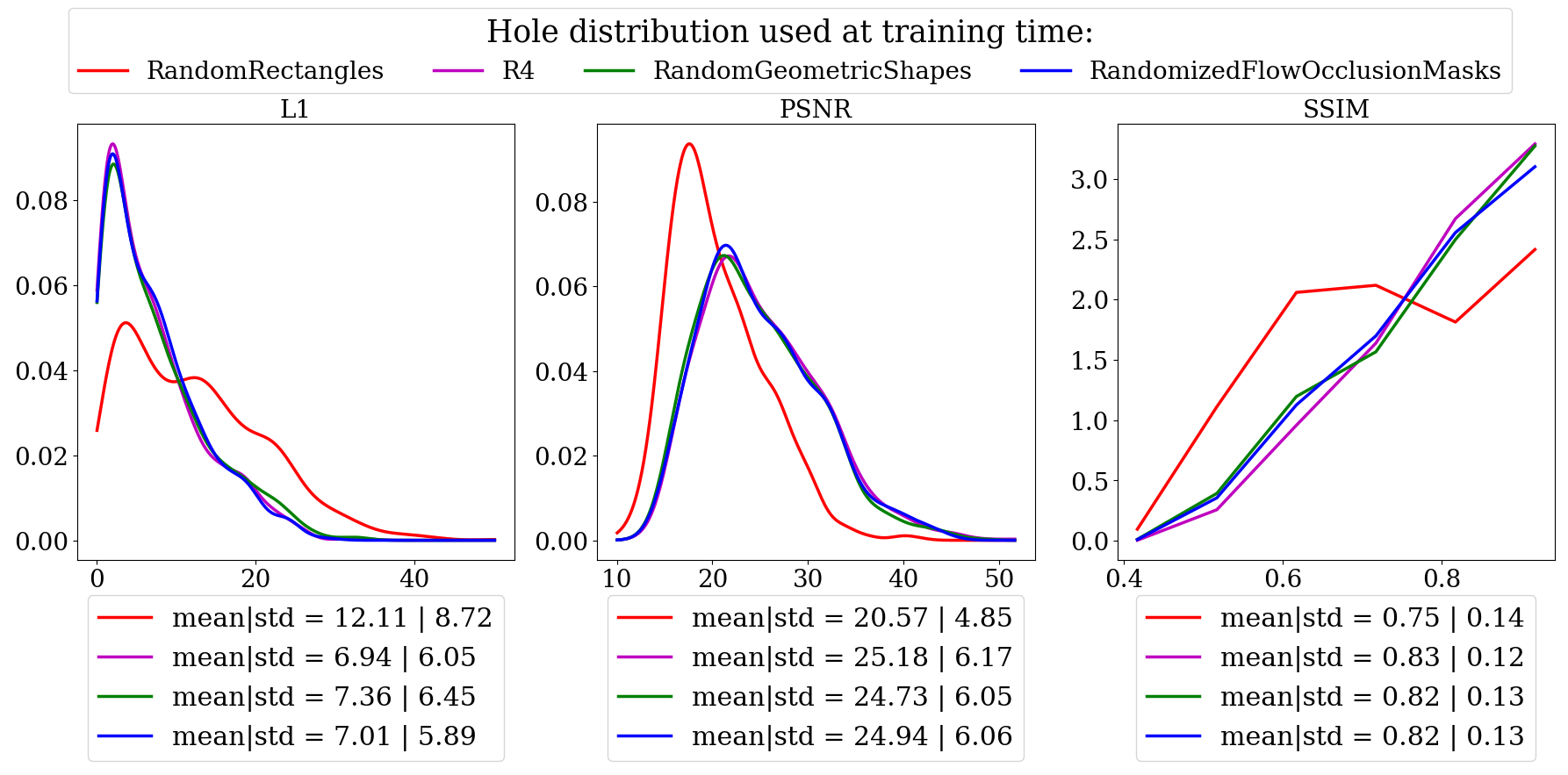}
    	\rulesep
        \includegraphics[width=0.5\linewidth]{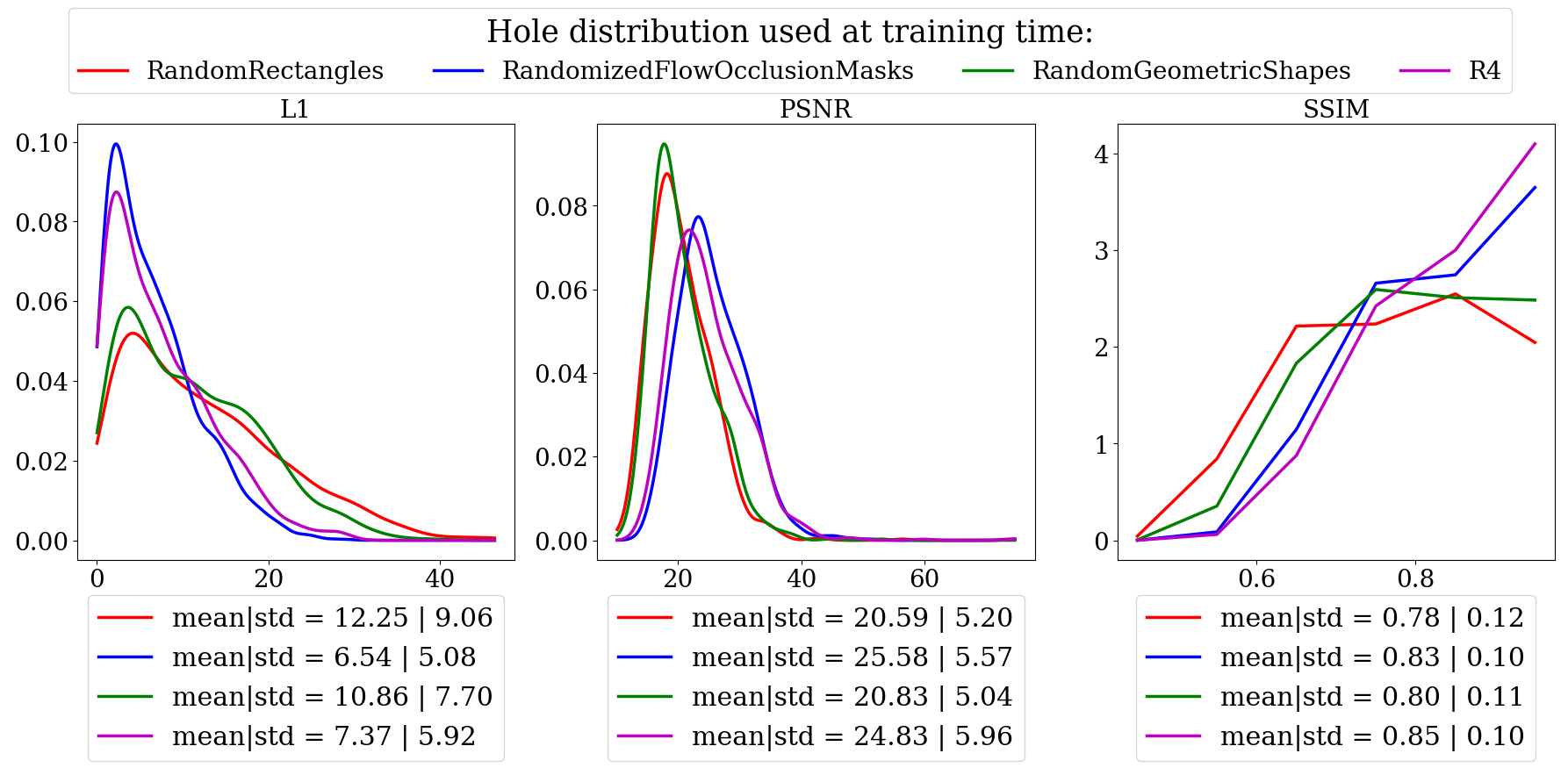}
        \label{pf_f32_rfom}
    }
    \caption{Smoothed empirical histograms of quantitative metrics evaluated over a test set, for a 10.3MB in-painting neural network trained on MIT Places\cite{NIPS2014Places} validation dataset (left column figures) and on CUHK CelebA \cite{liu2015faceattributes} 90/10 split dataset (right column figures) and tested with each of the four hole configurations. Training occurred for 2+1 epochs---at constant and at exponentially decaying learning rate. \textbf{Colors:} \textit{Red} represents training with $RR$, \textit{magenta} with $R4$, \textit{green} with $RGS$, and \textit{blue} with $RFOM$.}
    \label{fig:pf_f32}
\end{figure*}

Figure \ref{fig:pf_f32} shows quality-metric histograms (for patched images) when using an in-painting network in a train-test grid of the type given in Table \ref{tab:means_table}. 
We have color-coded the smoothed quality metric histograms in Figure \ref{fig:pf_f32}, according to the training hole distribution used. Red represents $RR_{train}$, magenta $R4_{train}$, green $RGS_{train}$, and blue $RFOM_{train}$.
The quality results are obtained by testing over a set of randomly-selected 1024 images. 
We display the results for the MIT Places dataset on the left panel of Figure \ref{fig:pf_f32} and CUHK Celeb-A dataset on its right panel. 
Viewing the plots in Figure \ref{pf_f32_rr}, we observe that a network trained with a set of random rectangles (i.e., $RR_{train}$,  red lines) does better when tested with the same hole distributions ($RR_{test}$). It does worse when tested with the other hole distributions. (See red lines' shift with respect to the others in Figures \ref{pf_f32_rr} to \ref{pf_f32_rfom}.). This result is as expected:
\begin{hyp}[H\ref{hyp:self}] \label{hyp:self}
A network trained with a particular hole distribution will generally have a bias towards that hole distribution. 
\end{hyp}
Hypothesis H\ref{hyp:self} is well-understood.
Neural networks are input-output function estimators sensitive to the input priors. 
Given that hole distributions in production environments are generally hard to predict (unless of course, only some hole shapes are allowed and others are disallowed), studying robustness to variations in hole distribution becomes relevant.  

Next, the results in  Figures \ref{pf_f32_rr} to \ref{pf_f32_rfom} show that when our network is trained with $RR$ hole distribution, it does significantly worse when tested with $RGS$ and $RFOM$ hole distributions in cross comparisons. We hypothesized that 
\begin{hyp}[H\ref{hyp:kernel}] \label{hyp:kernel}
When hole edges are aligned with the convolution kernels, the network trained will be less robust with respect to changes in hole distribution.
\end{hyp}
To test this hypothesis H\ref{hyp:kernel}, we conducted experiments by randomly rotating $RR$ holes to obtain the $R4$ hole geometry distribution. We observed that $R4_{train}$ (magenta-colored graphs in Figure \ref{fig:pf_f32}) produced far better cross-comparison results than $RR_{train}$ networks. 
In fact, $R4_{train}$ competes in robustness with $RFOM_{train}$, i.e. the network trained with $RFOM$ hole distribution, which is reputed to be the best hole distribution for in-painting training\cite{liu2018partialinpainting}. This is clear in the closer overlap of magenta-colored histogram graphs of ($R4_{train}$, magenta) and ($RFOM_{train}$, blue). 

Furthermore, comparing the left (Places) and right (Celeb-A) panels in Figure \ref{fig:pf_f32} (i.e., reviewing the impact of the dataset on our train-test grid) suggests:  
\begin{hyp}[H\ref{hyp:datasets}] \label{hyp:datasets}
Image datasets differ in how well they reveal the differential in relative robustness of an in-painting network against varying hole distributions.
\end{hyp}
Hypothesis H\ref{hyp:datasets} is supported by the fact that metric histograms in the case of training (and testing) with Celeb-A dataset (results shown in the right panels in Figure \ref{fig:pf_f32}) have shifted with respect to each other more distinctly (i.e., better for relative robustness measurements) than when we used Places dataset (results shown in the left panels in Figure \ref{fig:pf_f32}). In short, Celeb-A dataset can better discriminate relative robustness in our cross-comparison grids. The right panel (Celeb-A) also shows the similarity in the robustness of $R4_{train}$ and $RFOM_{train}$, giving further support for hypothesis H\ref{hyp:kernel}. 

One explanation for this observation may be that when training with Celeb-A, the in-painting networks have a far better ``guided" in-painting task (a smaller set of relevant representation for shapes to learn). So, they may instead develop greater biases to the hole types used in training them.  

To investigate, further, what may be some other causes for hypothesis H\ref{hyp:datasets}, we observe that an important, central task in in-painting is the correct reconstruction of edges and that a dataset which offers a more evenly distributed set of edges to be crossed by random holes is more likely to be better as a cross-comparison, relative-robustness ``separator" dataset than a dataset that offers less evenly distributed high intensity edges. 
%

\begin{figure}[h]
    \centering
    \includegraphics[width=0.47\linewidth]{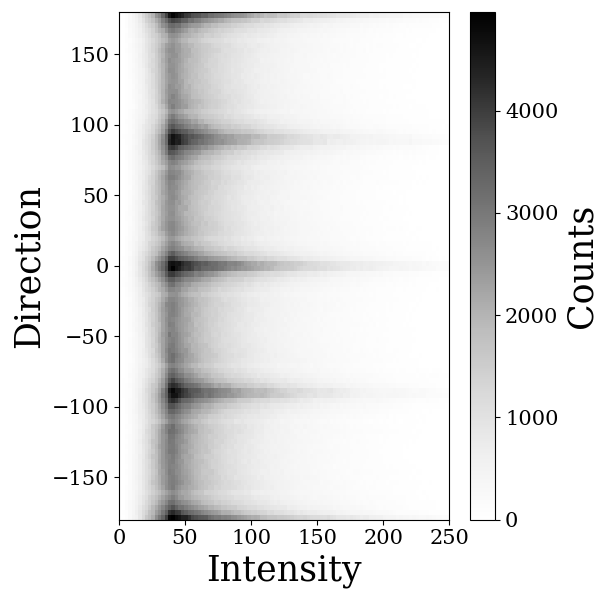}
    \rulesep
    \includegraphics[width=0.47\linewidth]{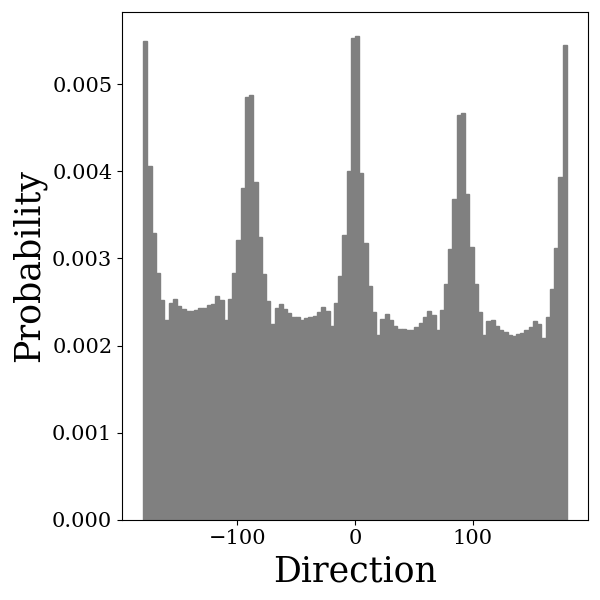}
    \\
    \includegraphics[width=0.47\linewidth]{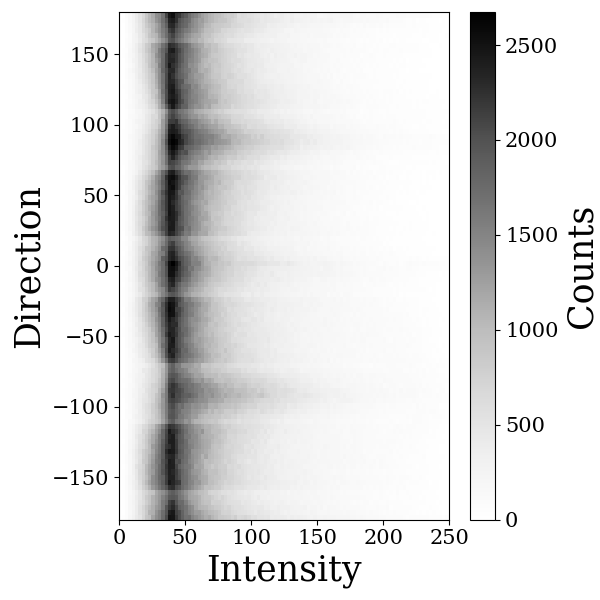}
    \rulesep
    \includegraphics[width=0.47\linewidth]{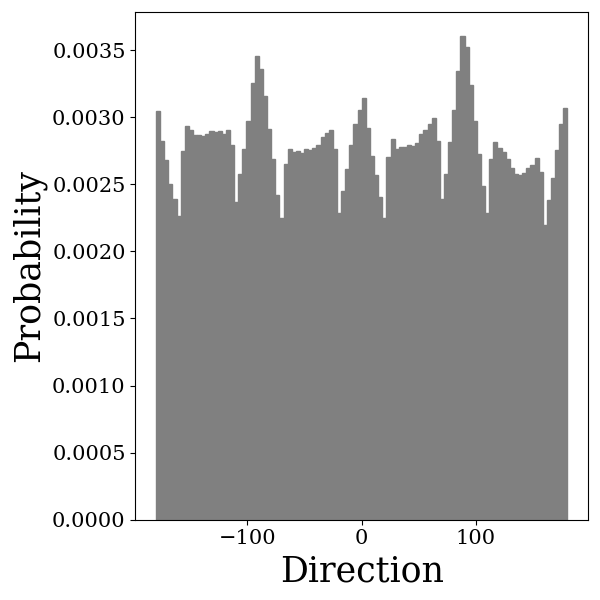}
    \caption{The heat maps on the left represent edge gradient PDFs $f(I,\zeta)$ of the edges in Places (top) and in Celeb-A (bottom) datasets. The edge gradient heat maps, on the right, provide the histogram for the computed gradients' intensity (horizontal) and direction (vertical). The charts on the right are 1-D PDFs of edge gradients, including only their direction.}
    \label{fig:dataset_edges}
\end{figure}

An examination of edge distribution using the Canny algorithm\cite{EdgeCanny1986} for the two datasets (see Figure \ref{fig:dataset_edges}) shows that the direction of edge gradients in the Places dataset have more pronounced modes, at 0, 90, 180 and 270 degrees, specially at higher levels of intensity. 
CelebA dataset, on the other hand, offers more evenly distributed edge gradient directions at similar range of intensities. 
This may provide another reason (or clue) for hypothesis H\ref{hyp:datasets}, and may also hint at an even more specific hypothesis, for whose more complete test and support, we would need to experiment with a much larger number of distinct datasets:
\begin{hyp}[H\ref{hyp:dataset_edges}] \label{hyp:dataset_edges}
Image datasets with more evenly distributed edge gradients are more likely to have greater capability in revealing in-painting relative robustness (robustness) differentials with respect to variations in hole distribution.
\end{hyp}

A hypothesis like the one in H\ref{hyp:dataset_edges} lead one to think about a whole new direction in the investigation of in-painting networks, as feature extractors, and the configuration of optimal training holes for that purpose.  
In closing this section, we note that there are many more  hole distributions and distinct image datasets which can be examined beyond the ones we have examined here, in order to further test, evaluate and better settle any of these hypothesis from an empirical point of view. 
%

\subsection{Cross-comparisons based on means of metrics}
\begin{table}
\begin{center}
\resizebox{\columnwidth}{!}{
\begin{tabular}{|l|l|l|l|l|l|}
\hline
Places & & & & & \\
 & $RR_{test}$ & $R4_{test}$ &  $RGS_{test}$ & $RFOM_{test}$ & $\alpha_m * \mathcal{R}_m$ \\
\hline
$m=L_1$ & & & & & \\
$RR_{train}$ & 3.66 & 8.68 & 13.4 & 12.1 &  0.40 \\
$R4_{train}$ & 3.7 &  3.47 & 5.78 & 6.94 & 0.87 \\
$RGS_{train}$ & 4.03 & 3.74 & 5.92 & 7.36 &  0.97 \\
$RFOM_{train}$ & 4.08 & 3.7 & 6.3 & 7.01 & 1 \\
\hline
$m=PSNR$ & & & & & \\
$RR_{train}$ & 26.1 & 20.8 & 18.8 & 20.6 & 0.79 \\
$R4_{train}$ & 25.9 & 26.5 & 24.8 & 25.2 &  0.999 \\
$RGS_{train}$ & 25.2 & 25.8 & 24.5 & 24.7 &  0.996 \\
$RFOM_{train}$ & 25.4 & 26.4 & 24.2 & 24.9 &  1 \\
\hline
$m=SSIM*10$ & & & & & \\
$RR_{train}$ & 8.89 & 8.01 & 7.08 & 7.52 & 0.87 \\
$R4_{train}$ & 8.9 & 8.96 & 8.36 & 8.31 & 0.989 \\
$RGS_{train}$ & 8.8 & 8.87 & 8.32 & 8.19 & 1 \\
$RFOM_{train}$ & 8.76 & 8.87 & 8.17 & 8.2 & 0.998 \\
\hline
\hline
Celeb-A & & & & & \\
& $RR_{test}$ & $R4_{test}$ & $RGS_{test}$ & $RFOM_{test}$ & $\alpha_m * \mathcal{R}_m$ \\
\hline
$m=L_1$ & & & & & \\
$RR_{train}$ & 5.13 & 14.7 & 13.5 & 12.3 & 0.48 \\
$R4_{train}$ & 7.44 & 7.47 & 6.62 & 7.37 & 0.91 \\
$RGS_{train}$ & 11.6 & 12.2 & 10.8 & 10.9 & 0.56 \\
$RFOM_{train}$ & 6.84 & 6.79 & 5.97 & 6.54 & 1 \\
\hline
$m=PSNR$ & & & & & \\
$RR_{train}$ & 25.9 & 18.8 & 18.6 & 20.6 & 0.80 \\
$R4_{train}$ & 22.9 & 23 & 24.2 & 24.8 &  0.99 \\
$RGS_{train}$ & 19.1 & 18.7 & 19.8 & 20.8 & 0.81 \\
$RFOM_{train}$ & 23.7 & 23.9 & 25.1 & 25.6 & 1 \\
\hline
$m = SSIM*10$ & & & & & \\
$RR_{train}$ & 8.6 &  7.21 & 7.18 & 7.77 &  0.89 \\
$R4_{train}$ & 8.22 & 8.22 & 8.3 & 8.45 & 1 \\
$RGS_{train}$ & 7.6 & 7.56 & 7.68 & 8 & 0.93 \\
$RFOM_{train}$ & 8.06 & 8.05 & 8.15 & 8.33 & 0.97 \\
\hline
\end{tabular}
}
\end{center}
\caption{Mean value of quality metrics in the train-test grid.}
\label{tab:means_table}
\end{table}

We evaluate the mean value of quality metric PDFs that we reviewed in Figure \ref{fig:pf_f32} and organized them in Table \ref{tab:means_table} in a train-test grid. 
Columns represent the test hole distributions, and the rows the training hole distributions, divided into sections for each of the three quality metrics we have computed. 
Note that the last column in the Table \ref{tab:means_table} provides the relative robustness, with respect to each quality metric type, normalized by the maximum observed for that particular quality metric. 
A value of 1 represents the best observed quality (in the mean) with respect to the corresponding metric. 

We first note that the trends in Figures \ref{fig:pf_f32} can also be confirmed---in a summary format---when comparing the means of the quality probability distributions, as shown in Table \ref{tab:means_table}.  
In fact, we used this finding in order to develop a (heuristic) "holistic" (expectation-based) measure of the relative in-painting robustness discussed in section \ref{sec:holistic}.  

Table \ref{tab:means_table} also indicates that $L_1$ is a better metric for differentiating relative robustness.
The next best separator is probably $PSNR$, and the least useful seems to be $SSIM$.
The mean values of $PSNR$ and $SSIM$, as gathered in Table \ref{tab:means_table} are generally quite close for networks trained with $R4$, $RGS$ and $RFOM$.

Our experiments indicate that training with $RFOM$ hole configuration leads to more robust in-painting networks. This confirms the claim regarding use of irregular holes in \cite{liu2018partialinpainting}, whose occlusion/de-occlusion files we used to generate our $RFOM$ holes. 
However, our findings also indicate that much of this improvement is purely due to the random angles of holes' edges (as the closeness of $R4_{train}$ results with those of $RFOM_{train}$ results suggest). The actual occlusion/de-occlusion shapes matter less. 

There remains other areas that require further exploration. For example, as a regularizing technique, one can study the impact of combining hole types---or generating holes that do not conform with any of the distributions---when training with any hole distribution from among a set such distributions. We leave this for future work. 

\subsection{Holistic comparisons}

Table \ref{tab:holistic_table} shows the results when we compute a (heuristic) ``holistic" robustness metric according to Equation \ref{eq:holistic}. 

It uses the data from Table \ref{tab:means_table} for this purpose. 
This relative robustness metric summarizes our relative robustness results. 
Again, our $RFOM_{train}$ networks prove more robust, and this robustness is more discernible when training (and testing) them with Celeb-A as opposed to Places. 
The relatively good ``holistic" robustness obtained by the $R4_{train}$ network across the two datasets, is further clear support for H\ref{hyp:kernel}. 
The variation in the ability to discriminate relative robustness of these networks, across Places and Celeb-A datasets, again, supports H\ref{hyp:datasets} even when we combine less and more relevant metrics with the same weight, as we have done in Table \ref{tab:holistic_table}.

\begin{table}
\begin{center}
\scalebox{0.9}
{
\begin{tabular}{|l|c|c|}
\hline
 & $\mathcal{R}_{holistic}$ & $\mathcal{R}_{holistic}$ \\
 & (Places) &  (Celeb-A) \\
\hline
$RR_{train}$ & 2.1 & 2.17 \\
$R4_{train}$ & 2.86 & 2.90 \\
$RGS_{train}$ &  2.96 & 2.30 \\
$RFOM_{train}$ &  2.998 & 2.97 \\
\hline
\end{tabular}
}
\end{center}
\caption{``Holistic" relative robustness scores. See Equation \ref{eq:holistic}.}
\label{tab:holistic_table}
\end{table}

\section{Discussion} \label{discussion}

The empirical work reported in this paper also relates to network robustness and sensitivity analysis.  
It is clear that the methodologies (empirical and theoretical) for the investigation of network sensitivity are fast evolving.
Although specific perturbations have been studied earlier\cite{sensitivity2018novak, sensitivity2019shu, Sharif2018AdversarialGN, DeepFool2016, Hein2017FormalGO}, the main course of sensitivity analysis studies has been focused on classification and identification tasks, adversarial attacks, and parameter perturbations. It appears that the specific case of perturbation of geometric distribution of holes in in-painting tasks has not been investigated.
It is clear that the performance of in-painting networks is quite sensitive to the PDF of the holes presented to them during training for a given image dataset.
Obviously, one can imagine a very large set of potentially useful hole distributions. 
In fact, there may be some arguments (based on the findings of the present paper) that exploring the space of hole distributions even further can lead to a more ``robust" network. 

Developing the statistical characteristics of ``filling the blank" systems as the ``blank distribution" varies has some parallels with understanding how despite noisy perceptual signals, every-day functional tasks are still performed quite easily.
For example, it is well-known, that in hearing we tend to fill gaps, quite readily and easily \cite{speech2016Chang}. 

A non-trivial general question worth exploring is the following: How does the statistical distribution of those gaps or holes affect a (``filling the blank") neural network's training or robustness? The method described in section \ref{sec:holistic} can be used in other situations, e.g., when the training set has a separable aspect (like the hole channel) whose statistics can be varied (``perturbed") independent of the statistics of images used and independent of the requirements of the task. It would be interesting to try these methods when investigating other ``filling the blank" tasks, e.g., as in \cite{FedusGD18}. 
Finally, note that in the case of partial convolutions, the presentation of the hole completely predetermines the form of the in-painting function applied. In a sense, with partial convolutions, the hole becomes a very specific function selector from a very large set of functions learned. The trainable parameters influence how the function then uses the context for that very specific hole.

\section{Conclusion and Future Work} \label{conclusion}

We have attempted to use in-painting task, subject to variations (``perturbation") of the (in-training) hole geometry distributions in order to take a step towards generalizing the concept of ``relative robustness'' of function estimation by means of neural networks in the special case of image in-painting.  
We have used cross-comparison, in a train-test grid, to study the robustness of an in-painting network trained (and cross-tested) under a variety of hole distributions.  

We would like to repeat our experiments with a far larger set of input ``hole" distributions and a larger variety of image datasets as well as open up a new perspective on the consequent network robustness and sensitivity analysis.

It would also be interesting to explore regularization techniques based on some mix of hole classes while training a neural network. It is not entirely clear, for example, whether the random introduction of some class of holes can help regularize training based on some other class of holes and what form that ``random introduction" needs to take.  

In yet another direction, it may be possible to adopt the perspective of adversarial robustness to analyze the impact of perturbations in hole geometry PDFs. What classes of hole patterns would produce particularly poor results for an in-painting system?  

We hope that in taking the modest step in this paper, we may have helped shed some more light on a potentially larger body of problems. We believe the more general problem is even more interesting. How do we characterize the relative robustness of networks when perturbations impact certain very controlled aspects in the training input---e.g., hole distribution for an in-painting network architecture when that network architecture makes no assumptions regarding the holes' geometry?

{\small
\bibliographystyle{ieee_fullname}
\bibliography{inpainting}
}
\clearpage
\pagebreak
\numberwithin{equation}{section}
\section{Supplementary Material}

Here, we add supporting material not included in the paper due to page constraints. 

\subsection{Training with Full Places Dataset} \label{larger_net}


\begin{figure*}[h]
    \centering
    \subfigure[Test w/ RandomRectangles - $RR$ hole configuration. $L_1$, $PSNR$, $SSIM$ histograms. Places. Left: Validation. Right: Full.] 
    {
        \includegraphics[width=0.5\linewidth]{RR_exp_3_sc_f32_places_4way}
    	\rulesep
        \includegraphics[width=0.5\linewidth]{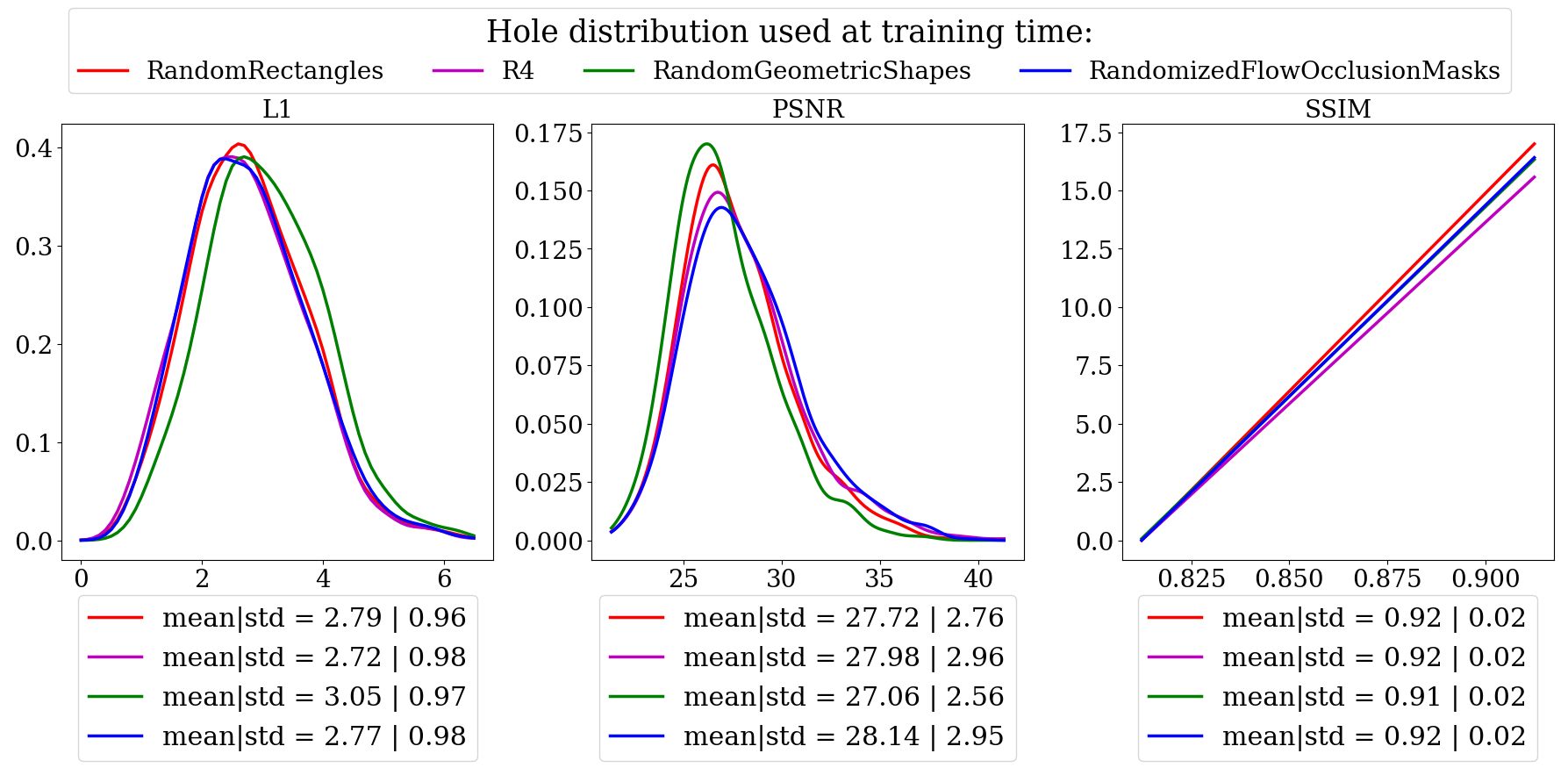}
        \label{fig:p_full_f32_rr}
    }
    \\
    \subfigure[Test w/ Randomly Rotated $RR$ - $R4$ hole configuration. $L_1$, $PSNR$, $SSIM$ histograms.  Places. Left: Validation. Right: Full.] 
    {
        \includegraphics[width=0.5\linewidth]{R4_exp_3_sc_f32_places_4way}
    	\rulesep
        \includegraphics[width=0.5\linewidth]{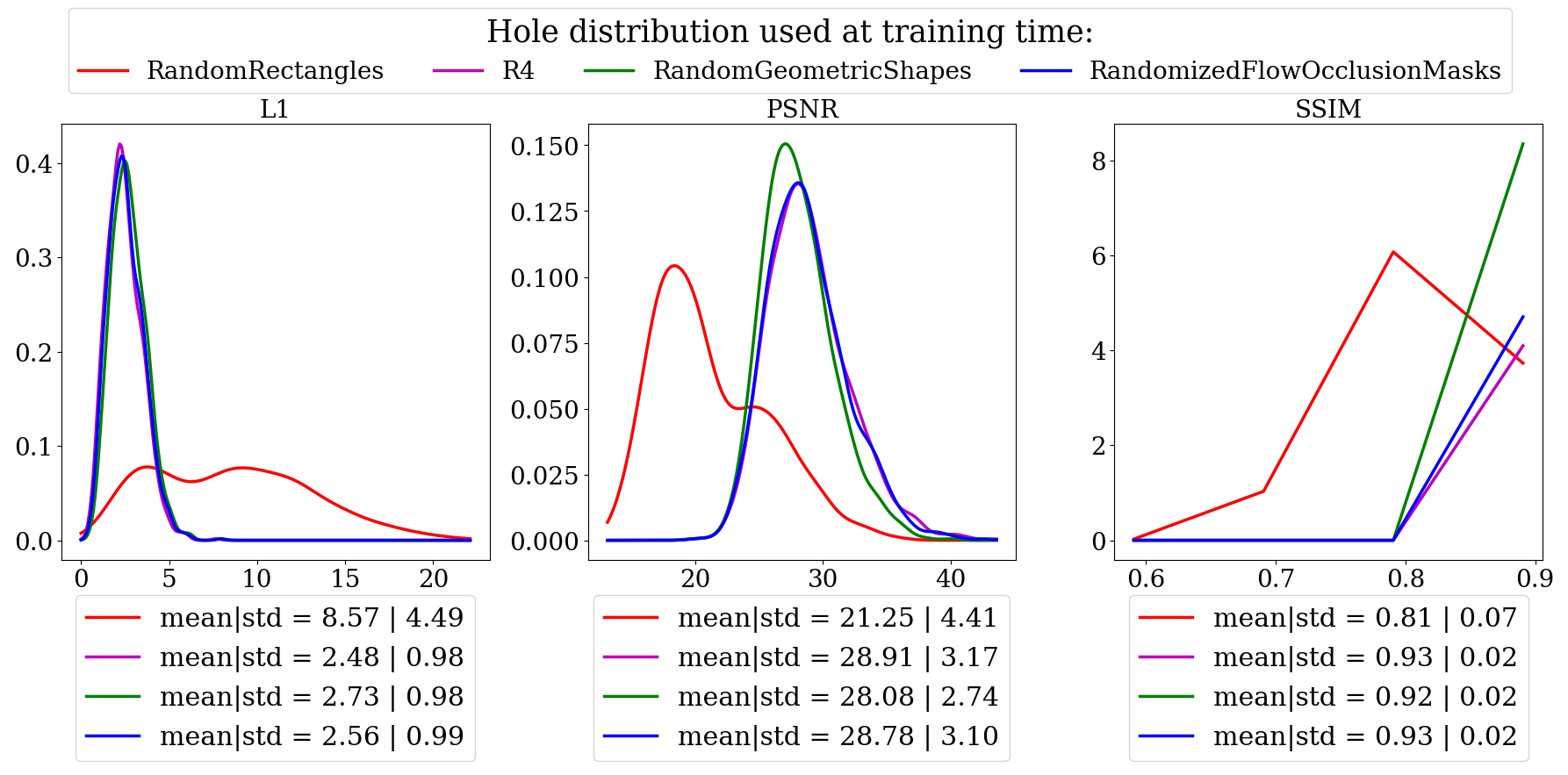}
        \label{fig:p_full_f32_r4}
    }
    \\
    \subfigure[Test w/ Random Geometric Shapes - $RGS$ hole configuration. $L_1$, $PSNR$, $SSIM$ histograms.  Places. Left: Validation. Right: Full.] 
    {
        \includegraphics[width=0.5\linewidth]{RGS_exp_3_sc_f32_places_4way}
    	\rulesep
        \includegraphics[width=0.5\linewidth]{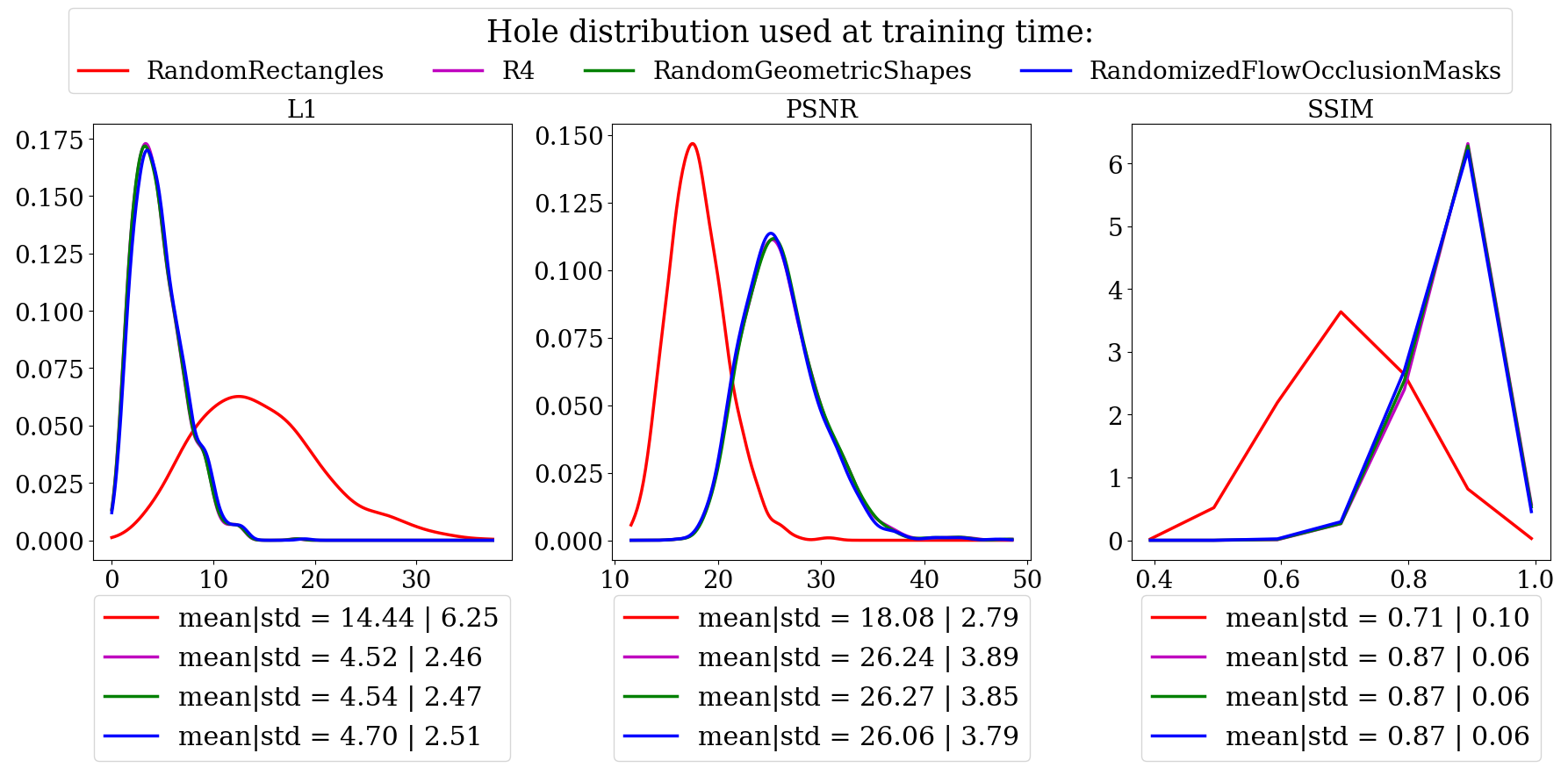}
        \label{fig:p_full_f32_rgs}
    }
    \\
    \subfigure[Test w/ Random Flow-based Occlusion/de-Occlusion Masks - $RFOM$ hole configuration. $L_1$, $PSNR$, $SSIM$ histograms.  Places. Left: Validation. Right: Full.] 
    {
        \includegraphics[width=0.5\linewidth]{RFOM_exp_3_sc_f32_places_4way}
    	\rulesep
        \includegraphics[width=0.5\linewidth]{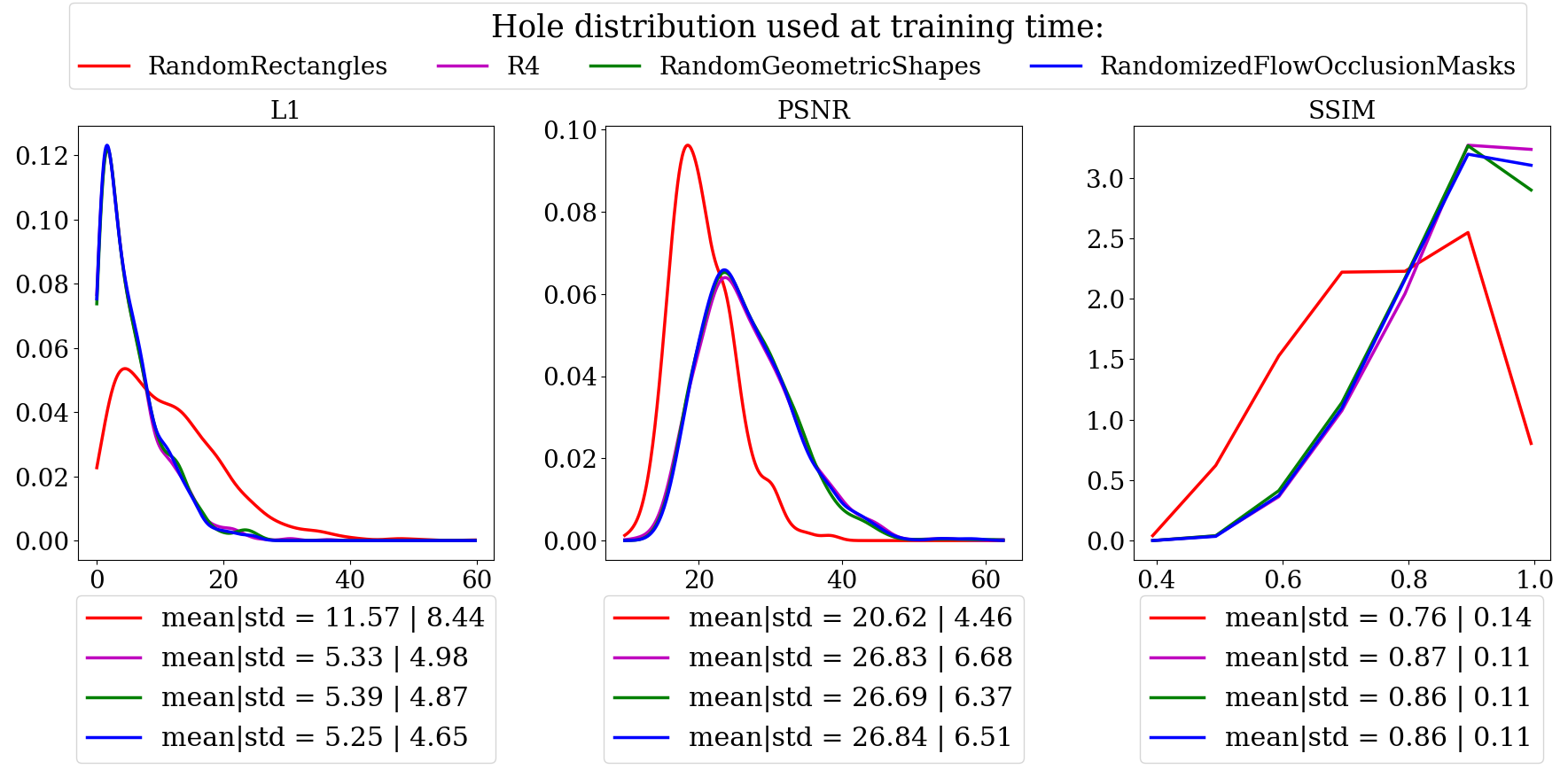}
        \label{fig:p_full_f32_rfom}
    }
    \caption{Smoothed empirical histograms of quantitative metrics evaluated in our train-test grid, for a 10.3MB in-painting neural network trained on MIT Places\cite{NIPS2014Places} validation dataset (\textit{left} column figures) and on MIT Places full dataset (\textit{right} column figures) and tested with each of the four hole configurations. Training occurred for 2+1 epochs---at constant and at exponentially decaying learning rate. \textbf{Colors:} \textit{Red} represents training with $RR$, \textit{magenta} with $R4$, \textit{green} with $RGS$, and \textit{blue} with $RFOM$.}
    \label{fig:p_full_f32}
\end{figure*}

MIT Places\cite{NIPS2014Places} validation dataset has roughly 36K images while the full dataset has 1.8M images. 

We state in the paper that using the full Places dataset validates our observation of relative-robustness trends and hypothesis. 

Figure \ref{fig:p_full_f32} shows that using the full MIT Places dataset gives further support to the observations in the paper and is consistent with the hypothesis there. 

In Figure \ref{fig:p_full_f32_rr}, we observe that the slight advantage that $RR_{train}$ had, when tested with $RR$ on the validation set, dissipates with longer training on the full set. 

However, the general trend for Hypothesis H\ref{hyp:self} holds when we review each of the rows of comparisons in Figure \ref{fig:p_full_f32}, i.e., when $i \neq r$, we can still observe that in most cases:

\begin{equation}\label{eq:self}
f_M(r,net_r) \succ f_M(r,net_i)
\end{equation}

In expression \ref{eq:self}, which represents Hypothesis H\ref{hyp:self}, $f_M(i,net_j)$ is the histogram (PDF) of some quality measure $M$---with some resilience functions potentially defined on its basis as in section \ref{sec:holistic}---when a network trained with hole configuration $j$ is tested against hole configuration $i$. 

The partial ordering symbol $\succ$ indicates dominance with respect to some quality measure, i.e., greater relative robustness with respect to that quality measure. 

In sections \ref{experimental} and \ref{results} of the paper, we have discussed some particular measure of dominance---in relative robustness. 

One measure of relative robustness was based on comparing histograms as in section \ref{results}, including holistic comparisons based on the statistical mean of the histograms, as described in section \ref{sec:holistic}.

A notation reminder: We generally use $r_{train}$ in the text to refer to $net_r$ in equations---a network trained with hole configuration $r$.

While longer training dissipates dominance of $RR_{train}$ over $H_{train}$, when tested on hole configuration $RR$ for any other hole configuration $H$, it does not negate Hypothesis H\ref{hyp:self}. 

That hypothesis, which is rooted in the basic relationship between a function estimation procedure and the data presented to it, is still validated. The validity of this hypothesis is, in fact, a requirement for the soundness of the statistical learning procedure used in all the cross-comparison cases we study. So, it is important to check H\ref{hyp:self}.

For the $2+1$ epochs of training, our results when training with the full Places dataset---as presented in Figures \ref{fig:p_full_f32_rr} to \ref{fig:p_full_f32_rfom}---still support Hypothesis H\ref{hyp:kernel}: 

\[ f_M(i,net_{RR}) \prec f_M(i,net_j) \]

for $i \neq RR$ and $j \neq RR$.

In fact, these results sharpen Hypothesis H\ref{hyp:kernel}, considering the vast improvement in robustness when a random rotation is added to $RR$ (obtaining $R4$). Figure \ref{fig:p_full_f32} shows how closely the results for $R4_{train}$ follow those for $RFOM_{train}$ and $RGS_{train}$, sometimes doing even better than the latter two. 

In general, the comparison of the right and left panels of Figures \ref{fig:p_full_f32_r4} to \ref{fig:p_full_f32_rfom} show the consistency of the trends observed in the paper---in the overlap and relative position of PDFs of the quality metrics over the test set. 

Longer training, over the full MIT Places dataset, neither negates the general trends nor causes any violation of any of the hypothesis we have stated in section \ref{results}.

\subsection{Train-Test Grid for a Larger Network}


\begin{figure*}[h]
    \centering
    \subfigure[Test w/ $RR$ hole configuration. $L_1$, $PSNR$, $SSIM$ histograms. Places Validation. Left: $N_{fm} = 32$ (10MB model). Right: $N_{fm} = 64$ (40MB model).]
    {
        \includegraphics[width=0.5\linewidth]{RR_exp_3_sc_f32_places_4way}
    	\rulesep
        \includegraphics[width=0.5\linewidth]{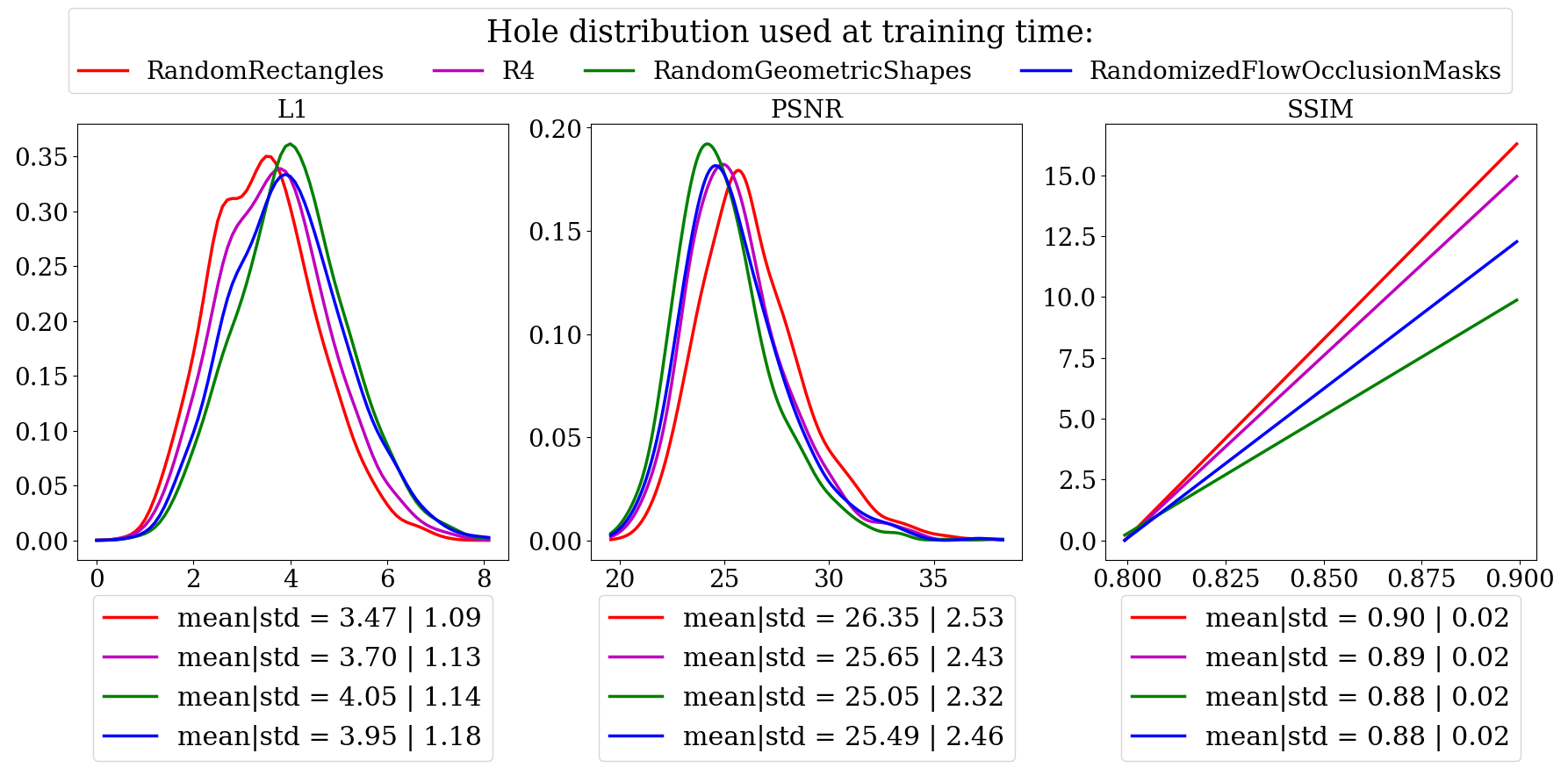}
        \label{fig:p3e3_f32_vs_f64_RR}
    }
    \\
    \subfigure[Test w/ $R4$ hole configuration. $L_1$, $PSNR$, $SSIM$ histograms. Places Validation. Left: $N_{fm} = 32$ (10MB model). Right: $N_{fm} = 64$ (40MB model).]
    {
        \includegraphics[width=0.5\linewidth]{R4_exp_3_sc_f32_places_4way}
    	\rulesep
        \includegraphics[width=0.5\linewidth]{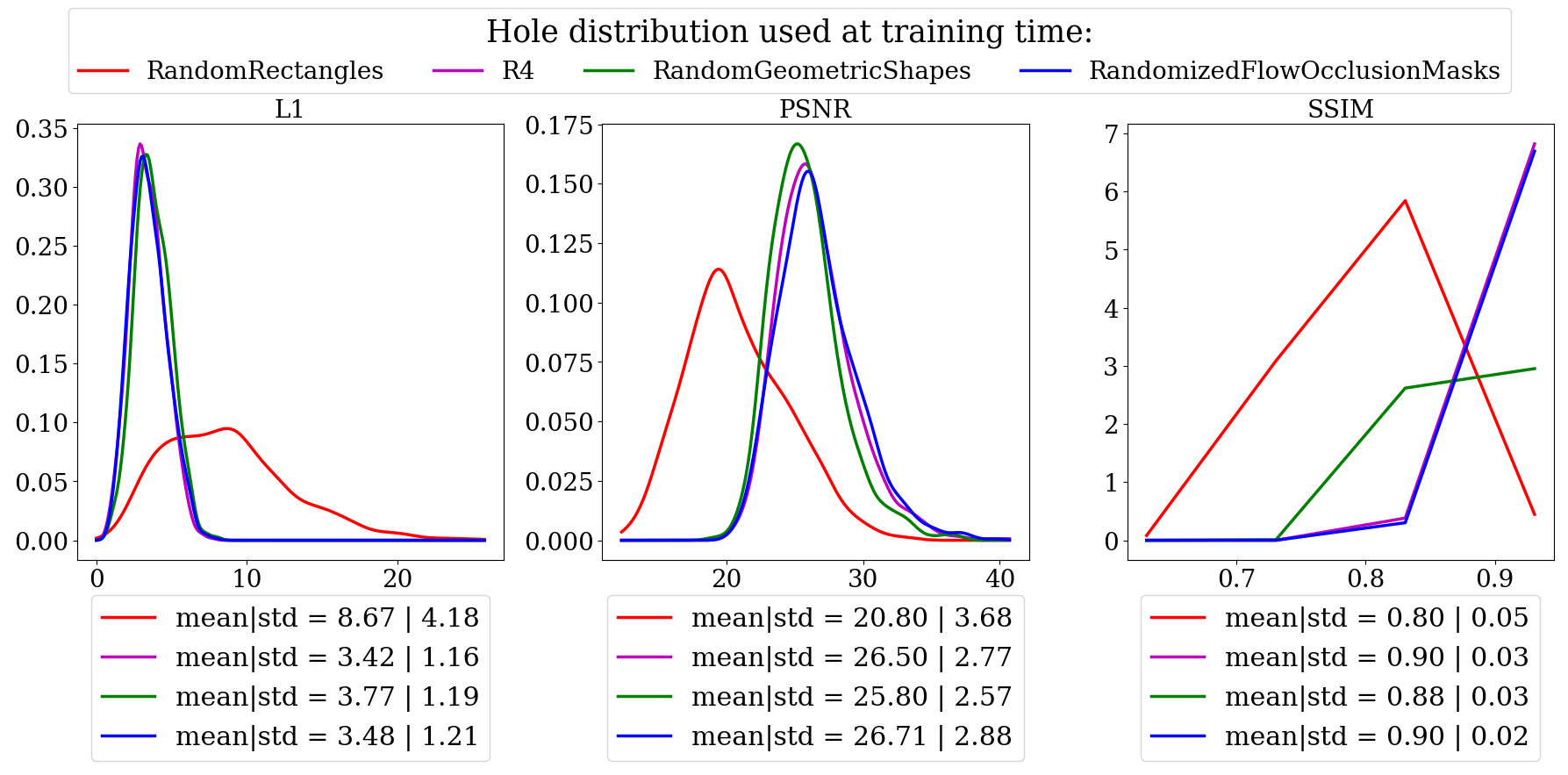}
        \label{fig:p3e3_f32_vs_f64_R4}
    }
    \\
    \subfigure[Test w/ $RGS$ hole configuration. $L_1$, $PSNR$, $SSIM$ histograms. Places Validation. Left: $N_{fm} = 32$ (10MB model). Right: $N_{fm} = 64$ (40MB model).]
    {
        \includegraphics[width=0.5\linewidth]{RGS_exp_3_sc_f32_places_4way}
    	\rulesep
        \includegraphics[width=0.5\linewidth]{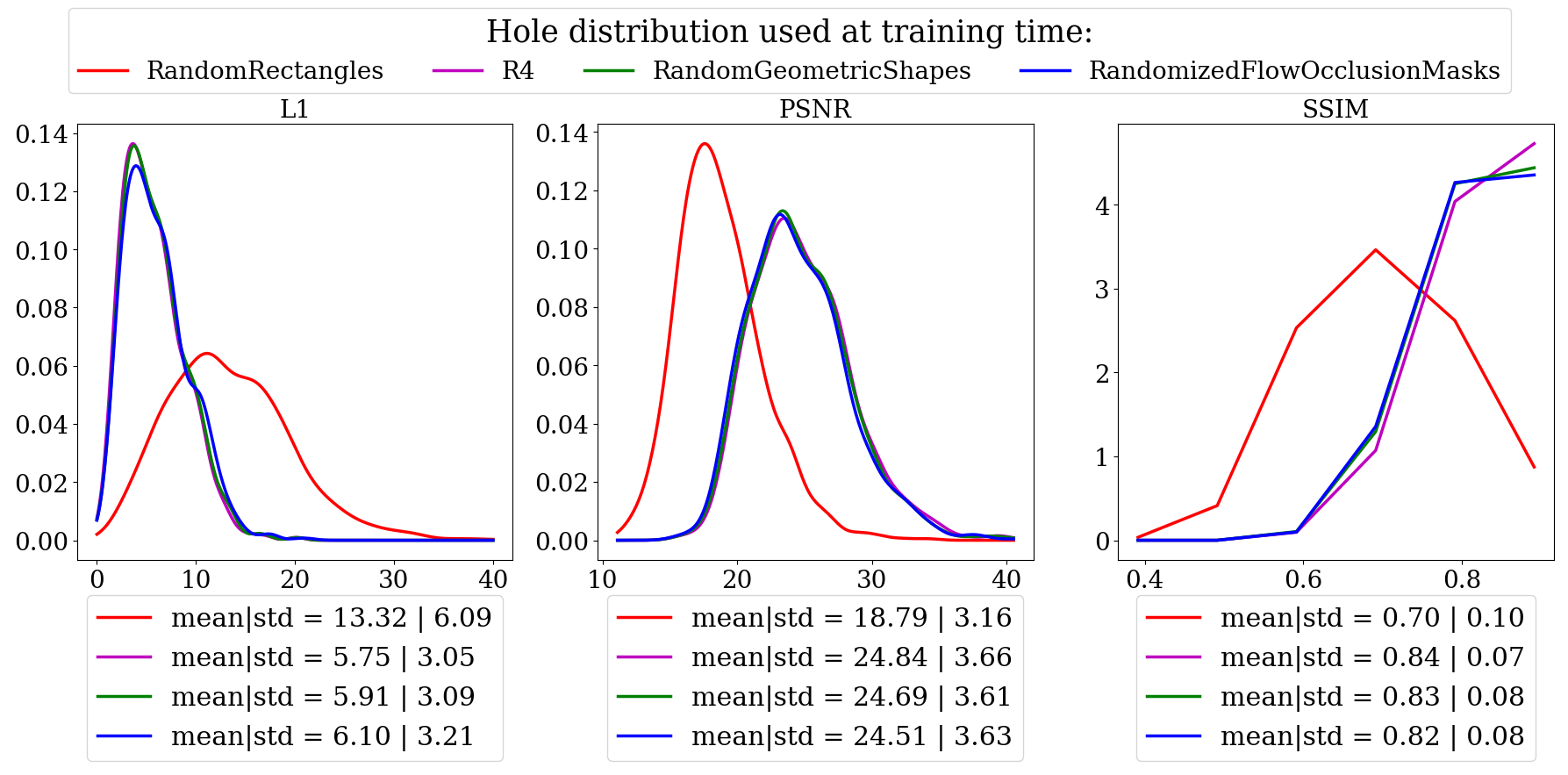}
        \label{fig:p3e3_f32_vs_f64_RGS}
    }
    \\
    \subfigure[Test w/ $RR$ hole configuration. $L_1$, $PSNR$, $SSIM$ histograms. Places Validation. Left: $N_{fm} = 32$ (10MB model). Right: $N_{fm} = 64$ (40MB model).]
    {
        \includegraphics[width=0.5\linewidth]{RFOM_exp_3_sc_f32_places_4way}
    	\rulesep
        \includegraphics[width=0.5\linewidth]{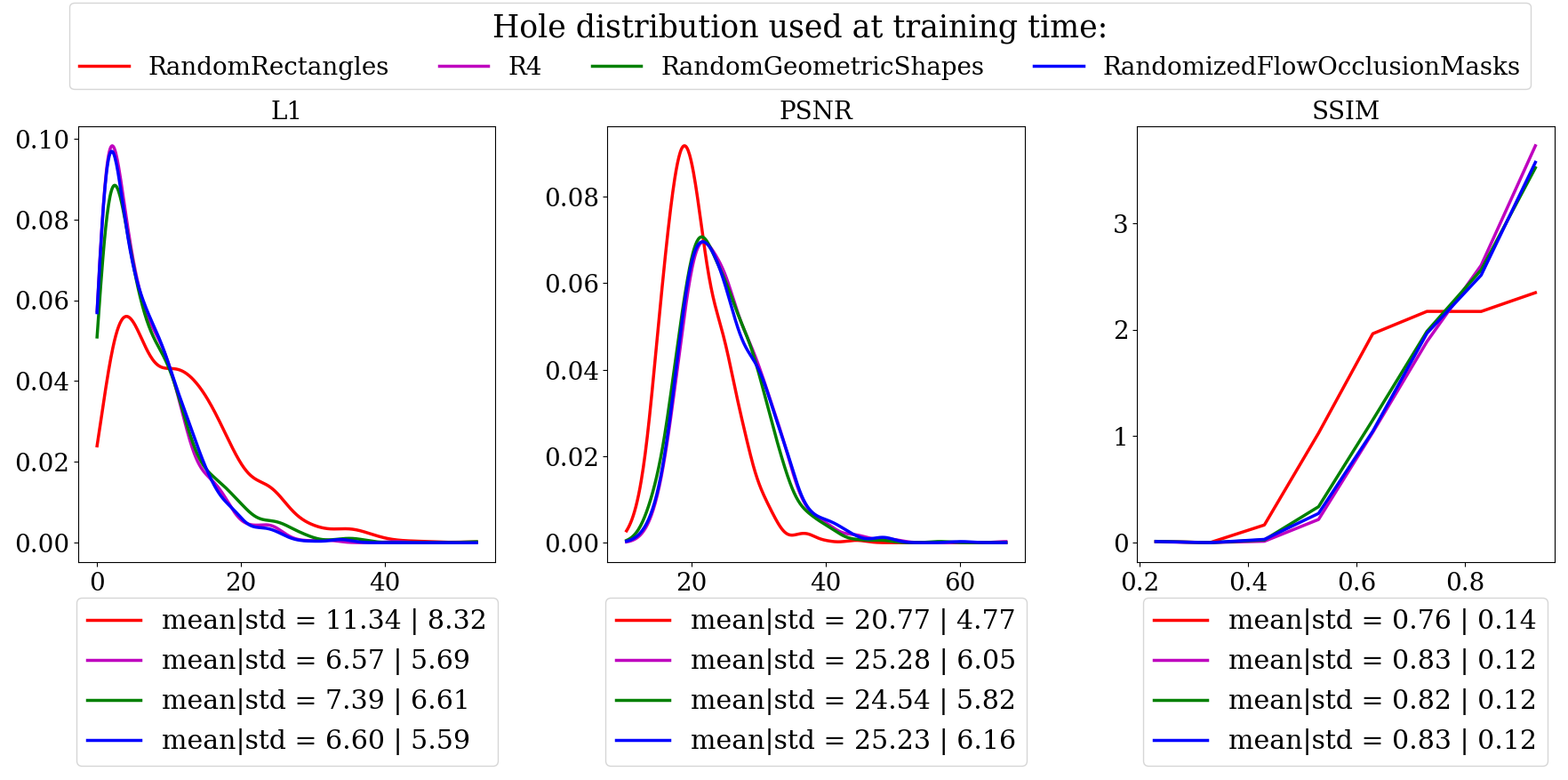}
        \label{fig:p3e3_f32_vs_f64_RFOM}
    }
    \caption{Smoothed empirical histograms of quantitative metrics evaluated over a test set. \textit{Left:} In-painting $N_{fm} = 32$ (\textbf{10MB}) neural network. \textit{Right:} In-painting $N_{fm} = 64$ (\textbf{40MB}) neural network. All trained on MIT Places\cite{NIPS2014Places} validation dataset using all hole configuration. Training occurred for 2+1 epochs---at constant and at exponentially decaying learning rate. \textbf{Colors:} \textit{Red} represents training with $RR$, \textit{magenta} with $R4$, \textit{green} with $RGS$, and \textit{blue} with $RFOM$.}
    \label{fig:p3e3_f32_vs_f64}
\end{figure*}

As mentioned in the paper, here we report our results obtained for a network with a larger capacity. 

Figure \ref{fig:p3e3_f32_vs_f64} shows our train-test grid for two cases---a small 10MB network and a larger 40MB network of the same architecture but a larger number of feature maps in each layer.  

The left panel of this figure is the same as the left panel of Figure \ref{fig:p_full_f32}. It shows the train-test grid results for the 10MB network, as described in the paper, where feature-map scaling factor is $N_{fm} = 32$ (see section \ref{experimental}). The right panel is for a 40MB network where the scaling factor is $N_{fm} = 64$. There are no other architectural differences between these networks other than the difference in their feature map capacity. Skip connections, contraction, kernel and dilated convolution patterns remain the same.

As Figure \ref{fig:p3e3_f32_vs_f64}, shows trends remain the same when we use a larger network that reported in the paper. We elide the details which would be along the lines of section \ref{larger_net}.  

\subsection{Train-Test Grid for a Larger Network, Trained Longer}


\begin{figure*}[h]
    \centering
    \subfigure[Test w/ $RR$ hole configuration. $L_1$, $PSNR$, $SSIM$ histograms. Larger (40MB) network, $N_{fm} = 64$, on Places Validation. Left: 2+1 epochs. Right: 10+10 epochs.]
    {
        \includegraphics[width=0.5\linewidth]{supp/RR_P3e3f64}
    	\rulesep
        \includegraphics[width=0.5\linewidth]{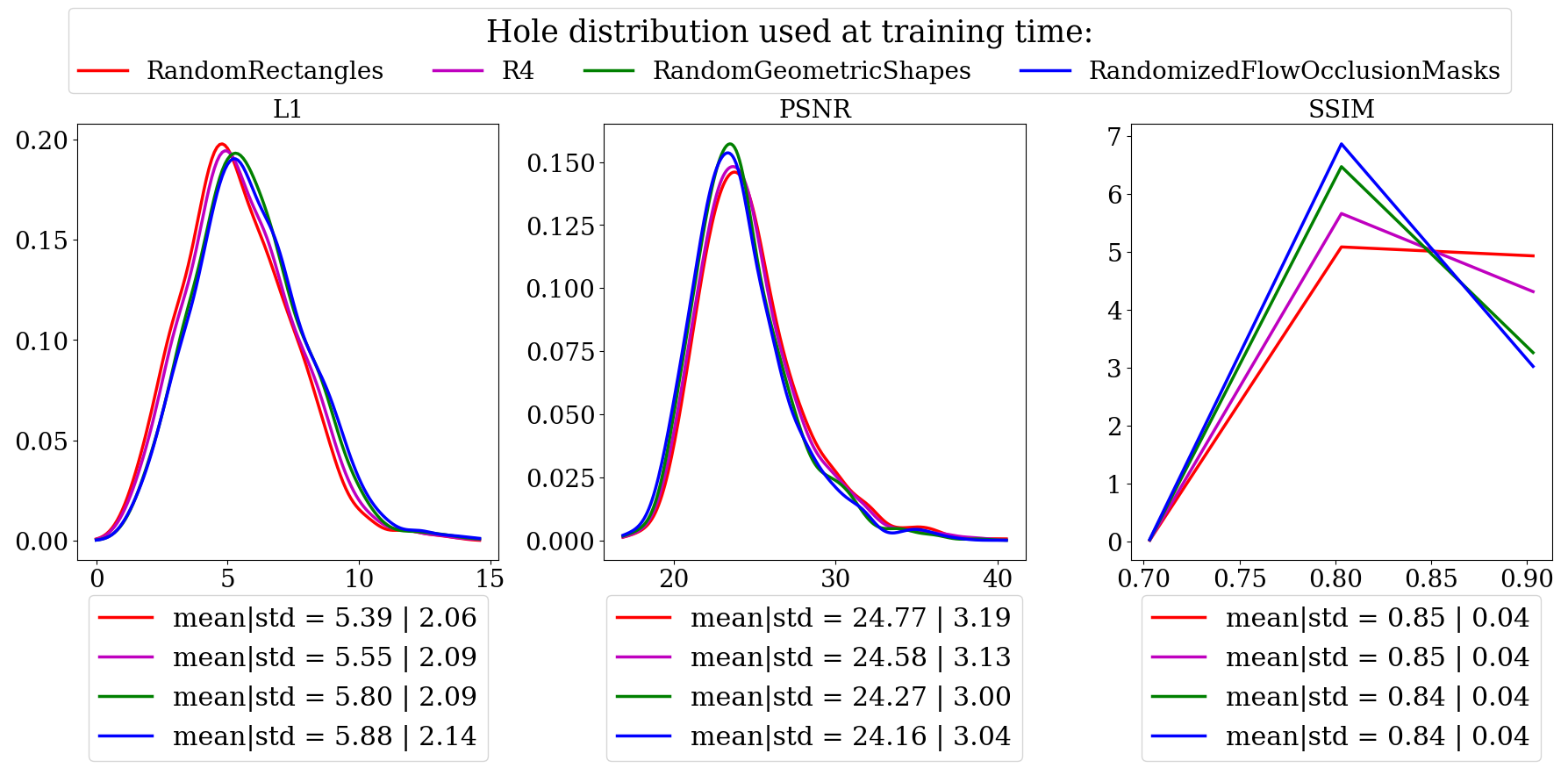}
        \label{fig:p20e3_f32_vs_f64_RR}
    }
    \\
    \subfigure[Test w/ $R4$ hole configuration. $L_1$, $PSNR$, $SSIM$ histograms. Larger (40MB) network, $N_{fm} = 64$, on Places Validation. Left: 2+1 epochs. Right: 10+10 epochs.]
    {
        \includegraphics[width=0.5\linewidth]{supp/R4_P3e3f64}
    	\rulesep
        \includegraphics[width=0.5\linewidth]{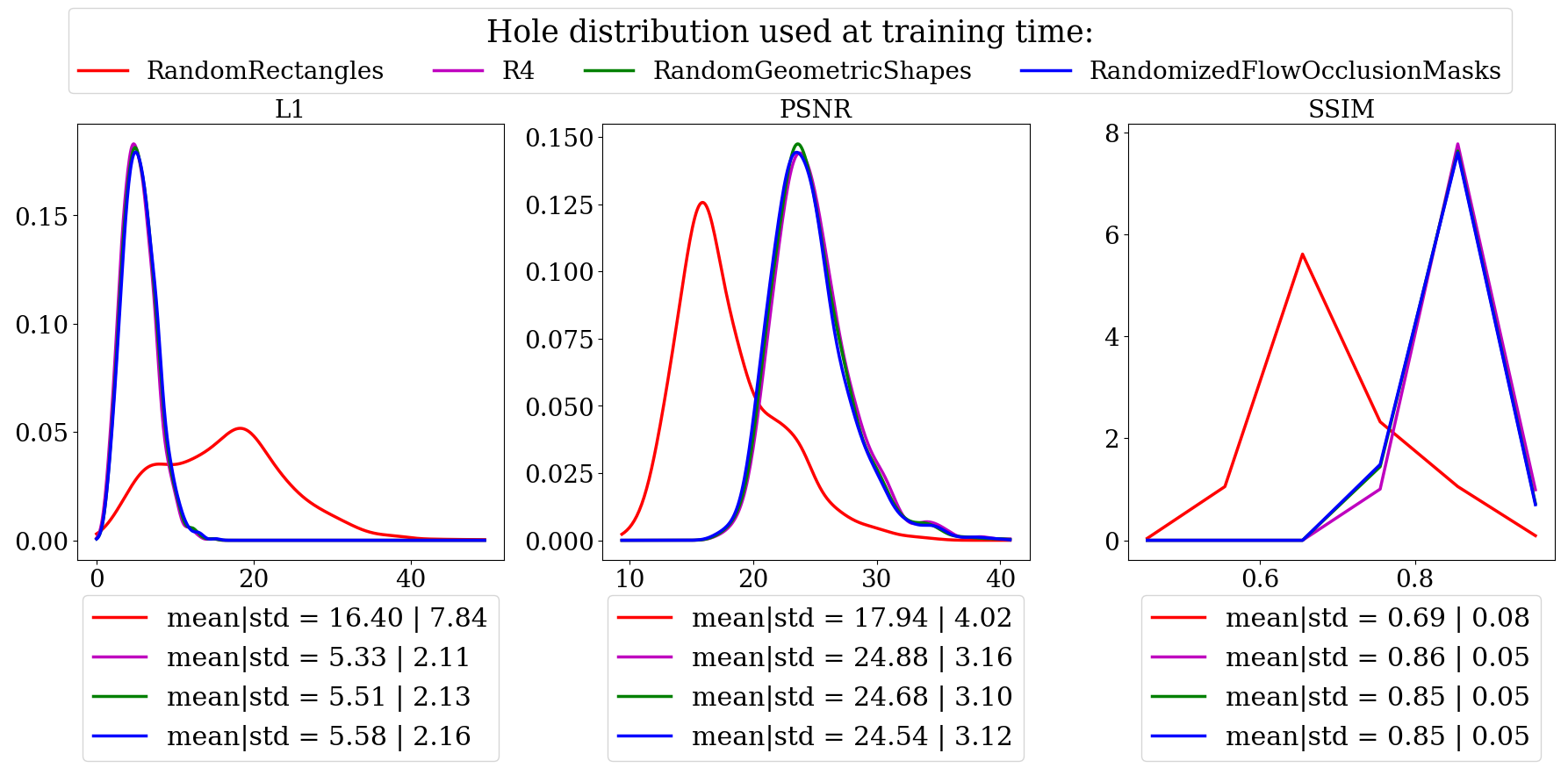}
        \label{fig:p20e3_f32_vs_f64_R4}
    }
    \\
    \subfigure[Test w/ $RGS$ hole configuration. $L_1$, $PSNR$, $SSIM$ histograms. Larger (40MB) network, $N_{fm} = 64$, on Places Validation. Left: 2+1 epochs. Right: 10+10 epochs.]
    {
        \includegraphics[width=0.5\linewidth]{supp/RGS_P3e3f64}
    	\rulesep
        \includegraphics[width=0.5\linewidth]{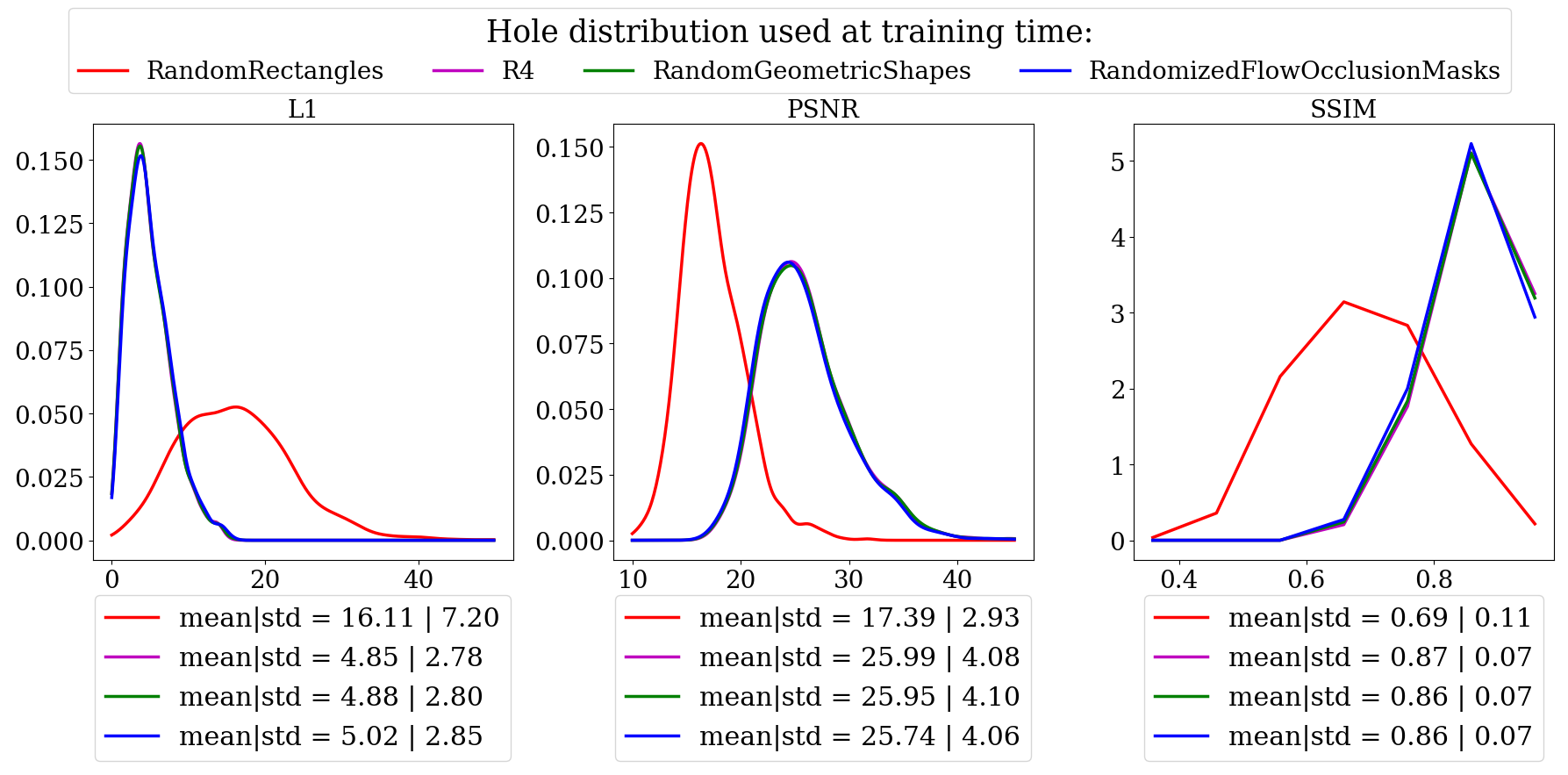}
        \label{fig:p20e3_f32_vs_f64_RGS}
    }
    \\
    \subfigure[Test w/ $RR$ hole configuration. $L_1$, $PSNR$, $SSIM$ histograms. Larger (40MB) network, $N_{fm} = 64$, on Places Validation. Left: 2+1 epochs. Right: 10+10 epochs.]
    {
        \includegraphics[width=0.5\linewidth]{supp/RFOM_P3e3f64}
    	\rulesep
        \includegraphics[width=0.5\linewidth]{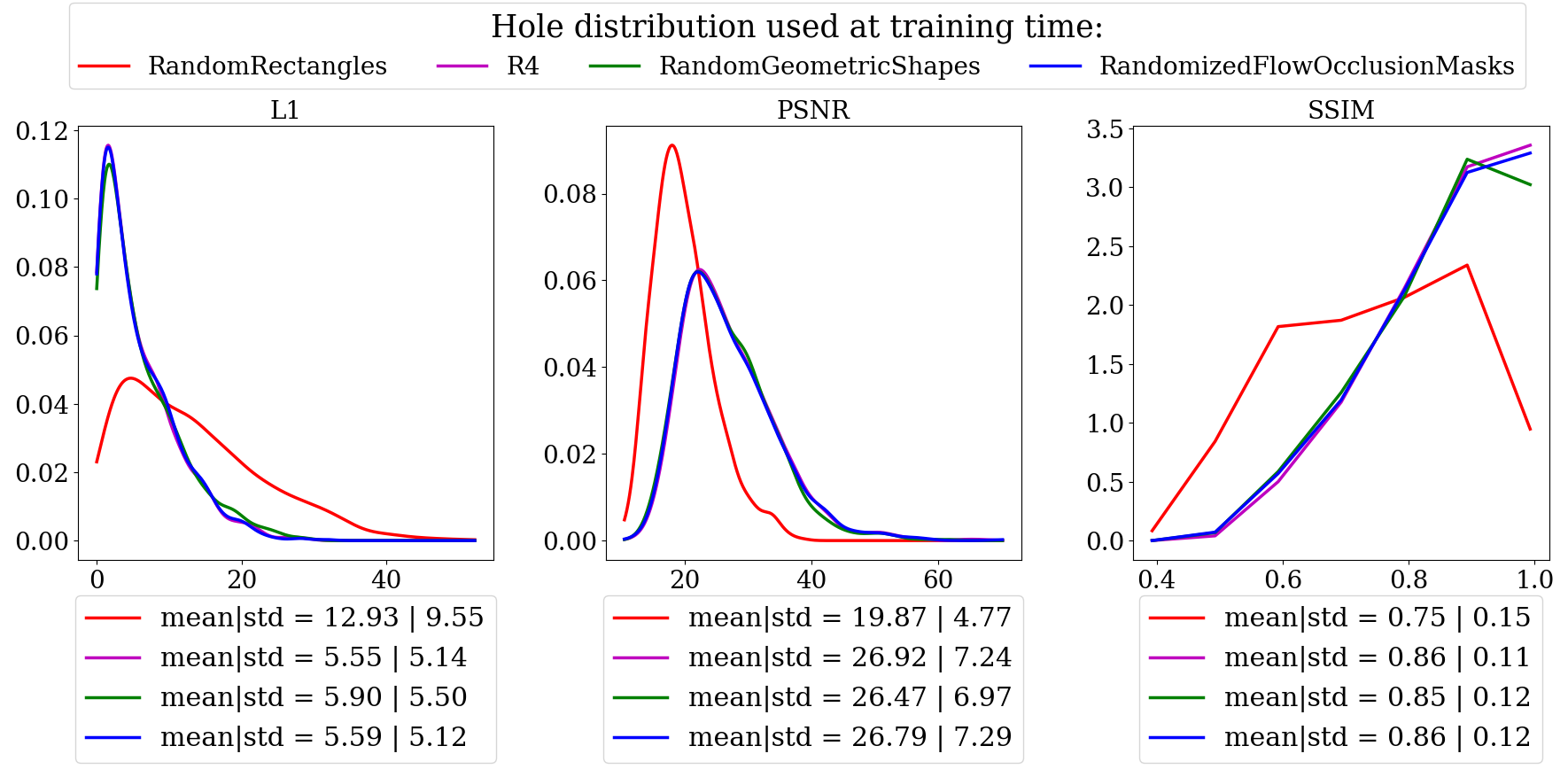}
        \label{fig:p20e3_f32_vs_f64_RFOM}
    }
    \caption{Smoothed empirical histograms of quantitative metrics evaluated over a test set. The larger (40MB) network, $N_{fm} = 64$, is trained on Places validation set. \textit{Left:} 2+1 epochs (as the main paper). \textit{Right:} 10+10 epochs of training. A $n+m$ epochs of training means $n$ epochs with constant learning rate and $m$ epochs with diminishing learning rate. \textbf{Colors:} \textit{Red} represents training with $RR$, \textit{magenta} with $R4$, \textit{green} with $RGS$, and \textit{blue} with $RFOM$.}
    \label{fig:p20e3_f32_vs_f64}
\end{figure*}

Figure \ref{fig:p20e3_f32_vs_f64}, shows that training the larger network longer than reported for cross-comparisons in the paper does not change the train-test relative robustness trends we have observed earlier. The comparison of the histograms in that figure should make this clear.  

\subsection{Datasets' Ability to Differentiate Relative Robustness}


\begin{figure*}[h]
    \centering
    \subfigure[Test w/ RandomRectangles - $RR$ hole configuration. $L_1$, $PSNR$, $SSIM$ histograms. Left: Places. Right: Celeb-A] 
    {
        \includegraphics[width=0.5\linewidth]{supp/RR_P3e3f64}
    	\rulesep
        \includegraphics[width=0.5\linewidth]{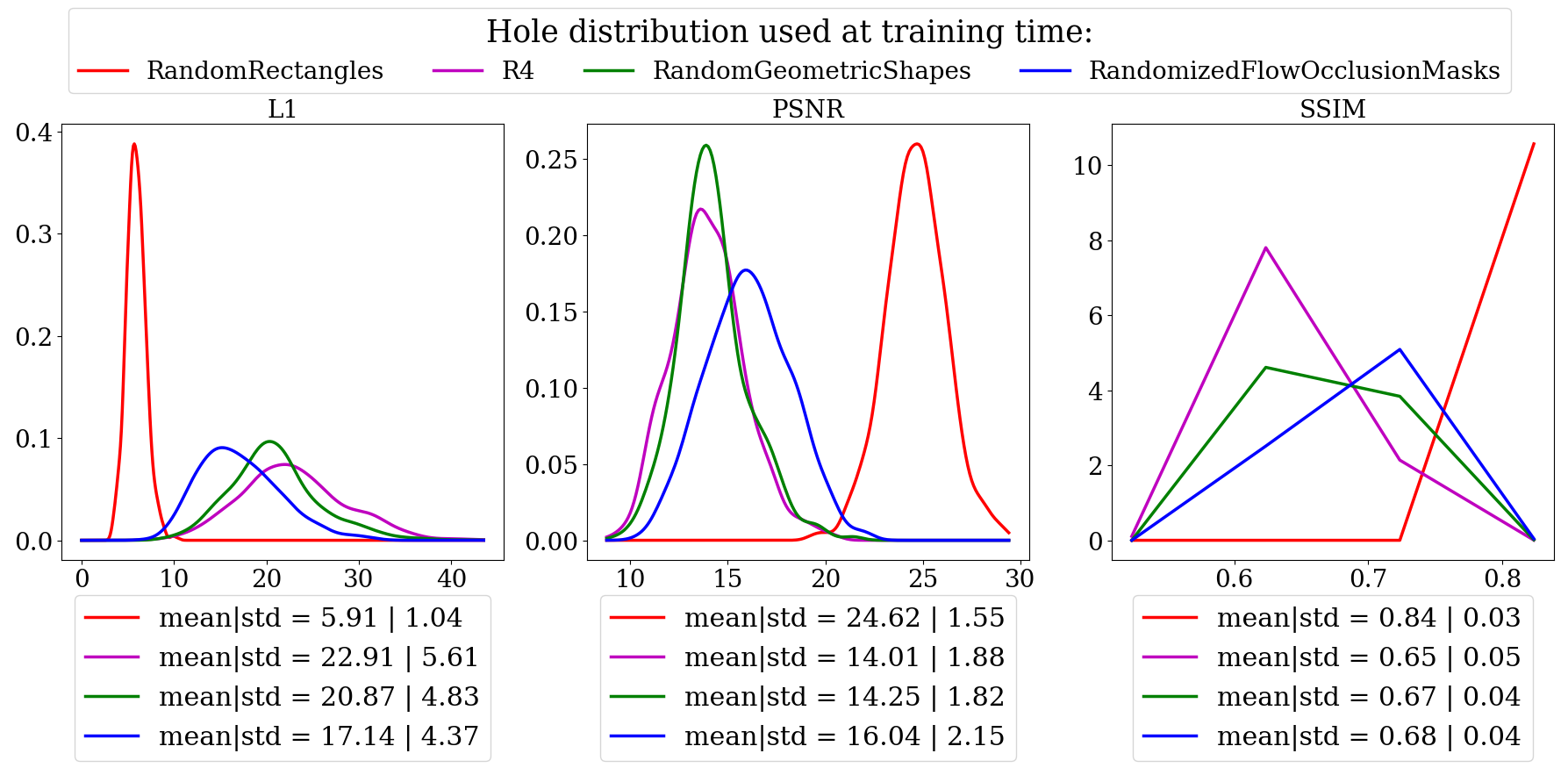}
        \label{pf_f64_rr}
    }
    \\
    \subfigure[Test w/ Randomly Rotated $RR$ - $R4$ hole configuration. $L_1$, $PSNR$, $SSIM$ histograms. Left: Places. Right: Celeb-A] 
    {
        \includegraphics[width=0.5\linewidth]{supp/R4_P3e3f64}
    	\rulesep
        \includegraphics[width=0.5\linewidth]{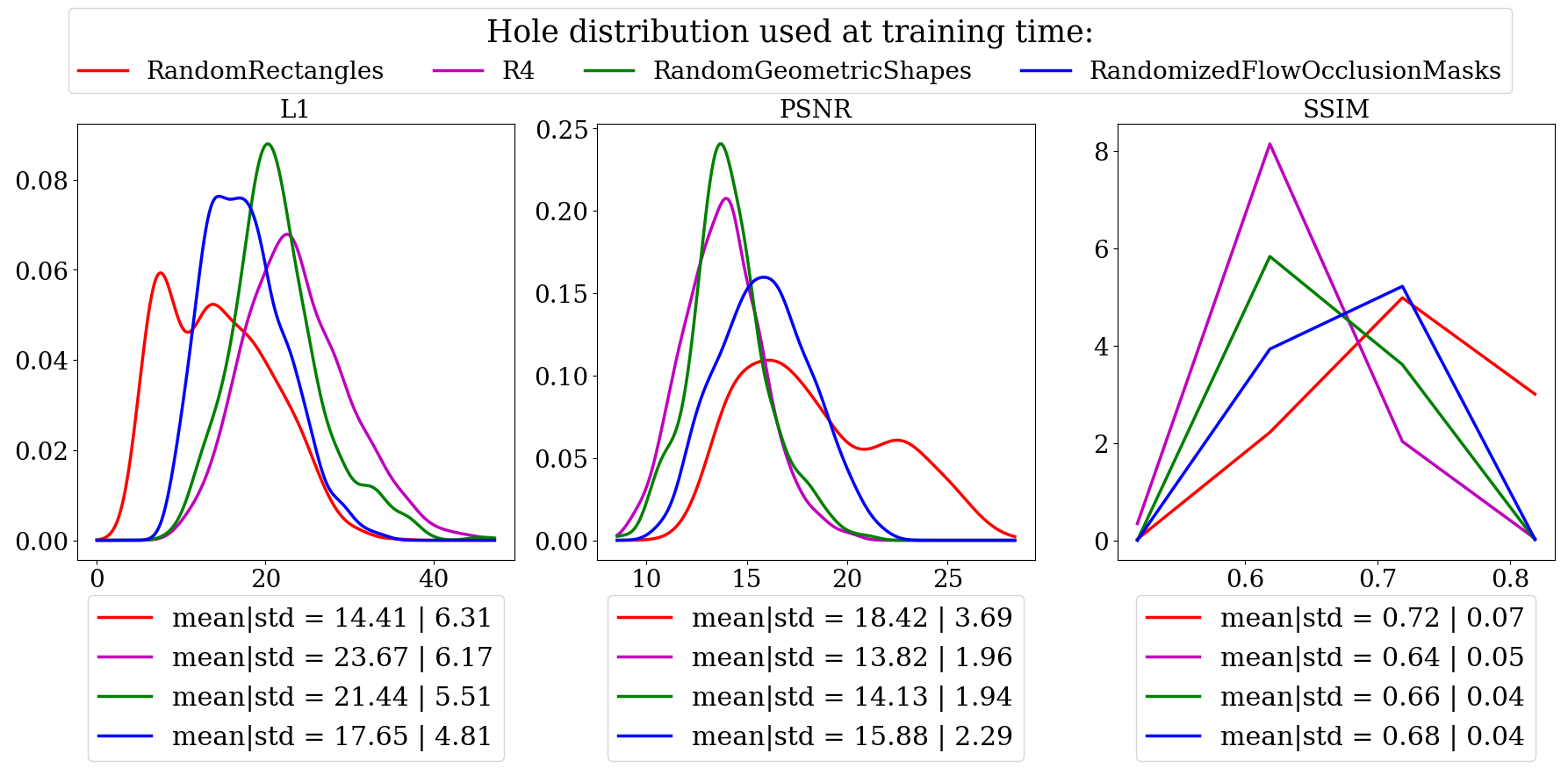}
        \label{pf_f64_r4}
    }
    \\
    \subfigure[Test w/ Random Geometric Shapes - $RGS$ hole configuration. $L_1$, $PSNR$, $SSIM$ histograms. Left: Places. Right: Celeb-A] 
    {
        \includegraphics[width=0.5\linewidth]{supp/RGS_P3e3f64}
    	\rulesep
        \includegraphics[width=0.5\linewidth]{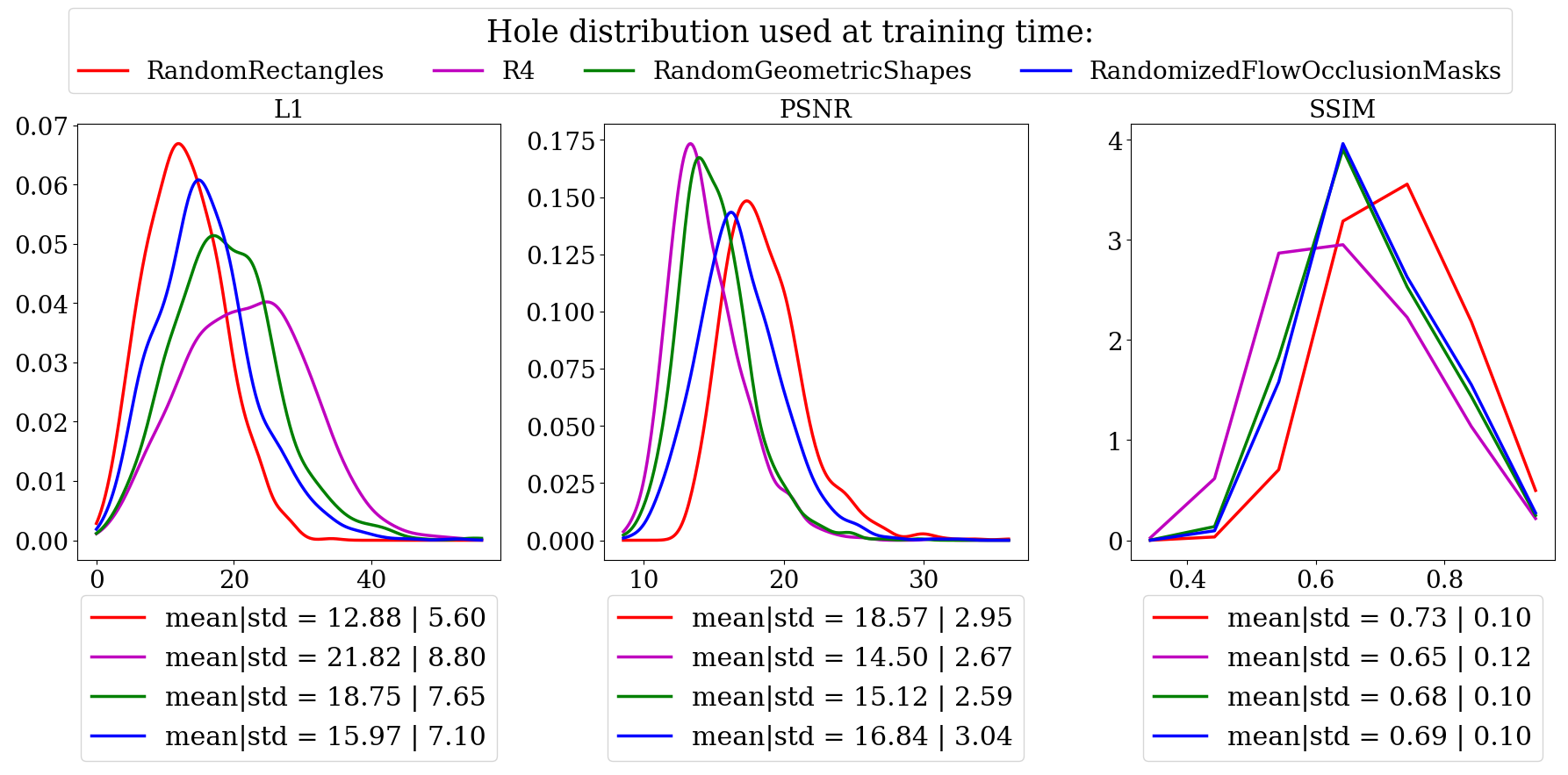}
        \label{pf_f64_rgs}
    }
    \\
    \subfigure[Test w/ Random Flow-based Occlusion/de-Occlusion Masks - $RFOM$ hole configuration. $L_1$, $PSNR$, $SSIM$ histograms. Left: Places. Right: Celeb-A] 
    {
        \includegraphics[width=0.5\linewidth]{supp/RFOM_P3e3f64}
    	\rulesep
        \includegraphics[width=0.5\linewidth]{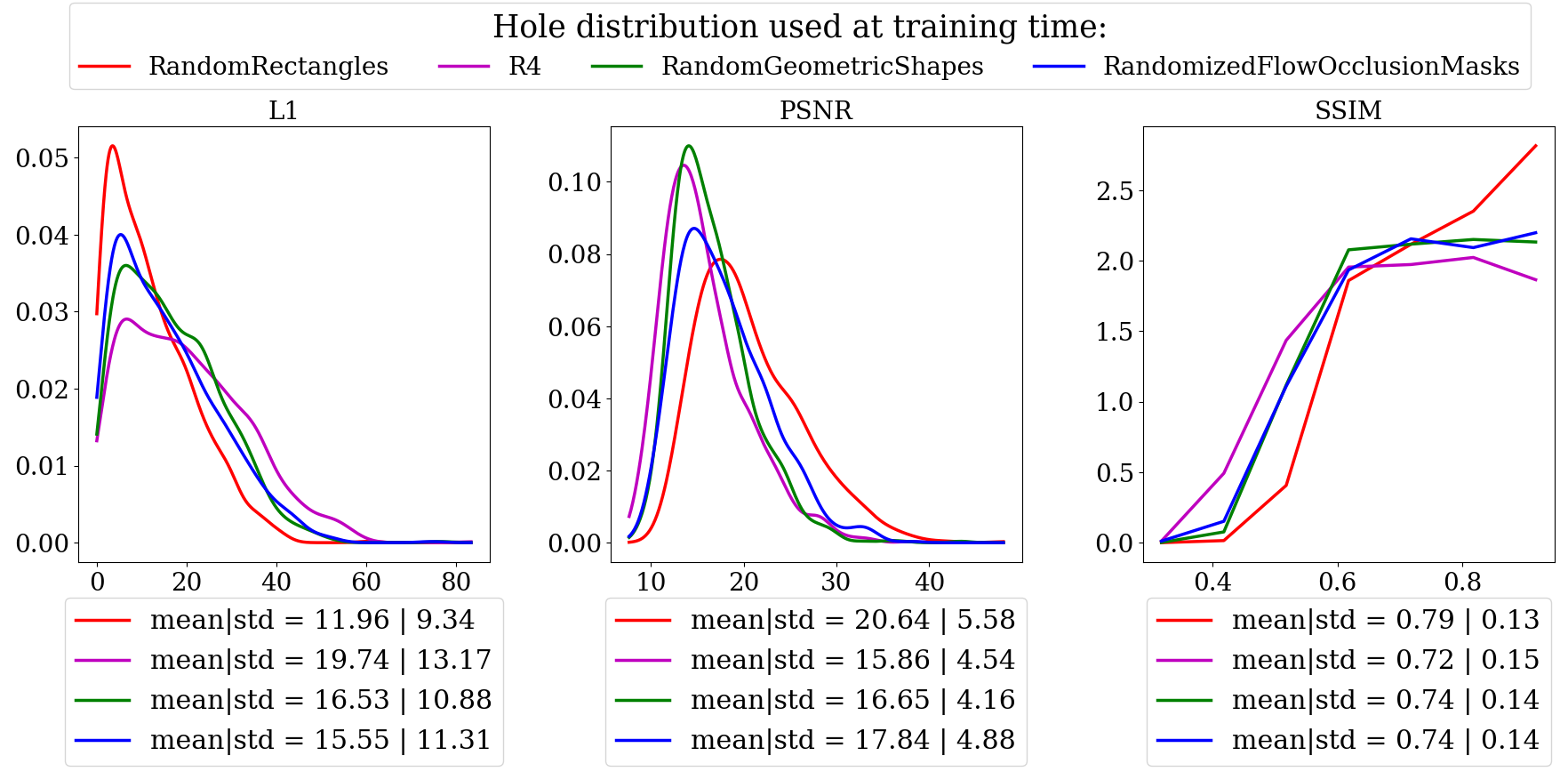}
        \label{pf_f64_rfom}
    }
    \caption{Smoothed empirical histograms of quantitative metrics evaluated over a test set, for a 40 MB ($N_{fm}=64$) in-painting neural network trained on MIT Places\cite{NIPS2014Places} validation dataset (\textit{left} column figures) and on CUHK CelebA \cite{liu2015faceattributes} 90/10 split dataset (\textit{right} column figures) and tested with each of the four hole configurations. Training occurred for 2+1 epochs---at constant and at exponentially decaying learning rate. \textbf{Colors:} \textit{Red} represents training with $RR$, \textit{magenta} with $R4$, \textit{green} with $RGS$, and \textit{blue} with $RFOM$.}
    \label{fig:pf_f64}
\end{figure*}

Finally, when  using larger models, the hypothesis H\ref{hyp:datasets} and H\ref{hyp:dataset_edges}, regarding variations in the ability of datasets to differentiate relative robustness of in-painting networks still holds. CelebA continues to be a better dataset for this differentiation even when we use larger models. See Figure \ref{fig:pf_f64}.

\subsection{Longer Training with CelebA}

\begin{figure*}[h]
    \centering
    \subfigure[Test w/ RandomRectangles - $RR$ hole configuration. $L_1$, $PSNR$, $SSIM$ histograms. Celeb-A. Left: 2+1 epochs. Right: 10+10 epochs.] 
    {
        \includegraphics[width=0.5\linewidth]{RR_exp_3_sc_f32_faces_4way}
    	\rulesep
        \includegraphics[width=0.5\linewidth]{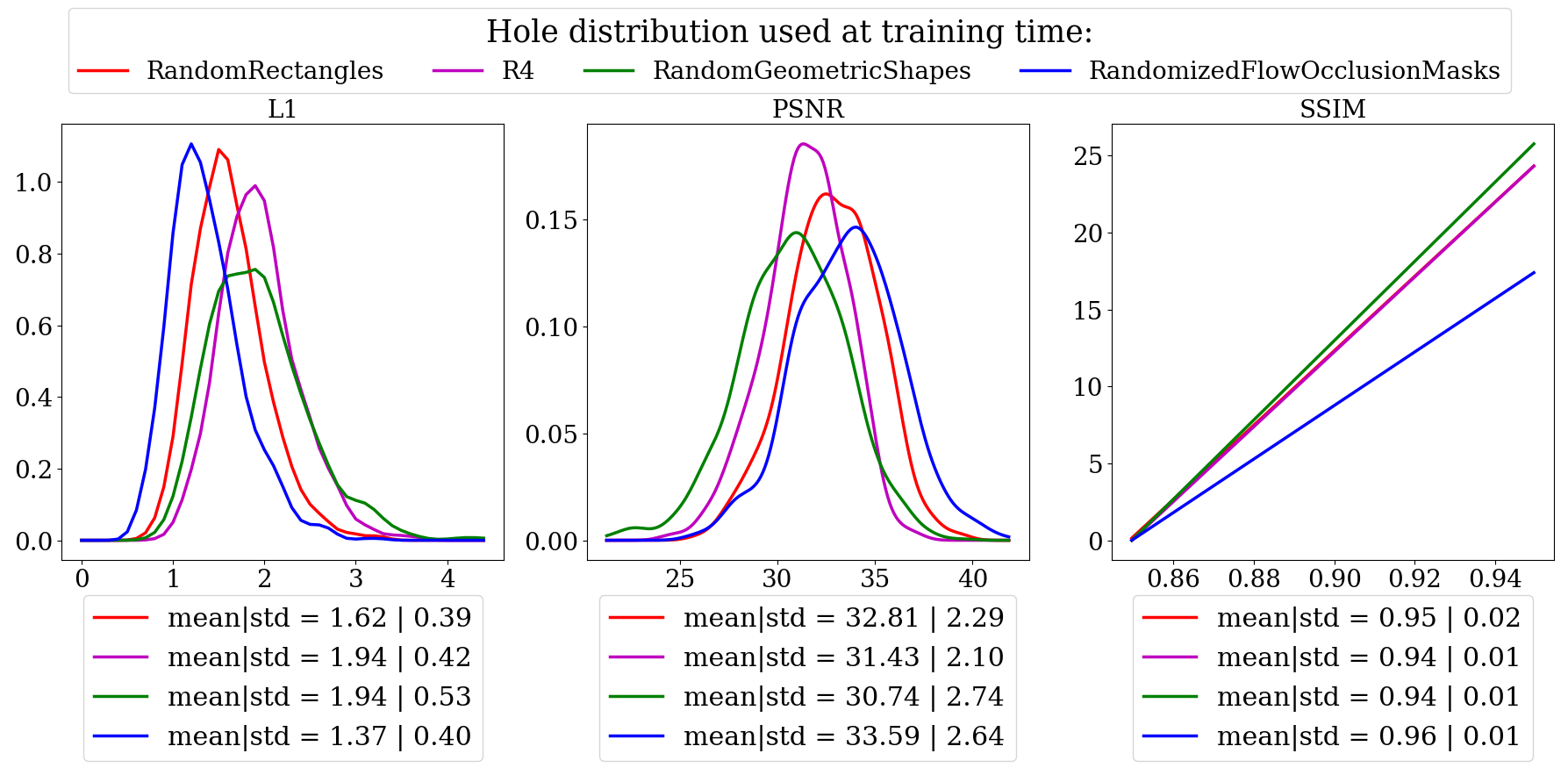}
        \label{F3_v_F20_rr}
    }
    \\
    \subfigure[Test w/ Randomly Rotated $RR$ - $R4$ hole configuration. $L_1$, $PSNR$, $SSIM$ histograms. Celeb-A. Left: 2+1 epochs. Right: 10+10 epochs.] 
    {
        \includegraphics[width=0.5\linewidth]{R4_exp_3_sc_f32_faces_4way}
    	\rulesep
        \includegraphics[width=0.5\linewidth]{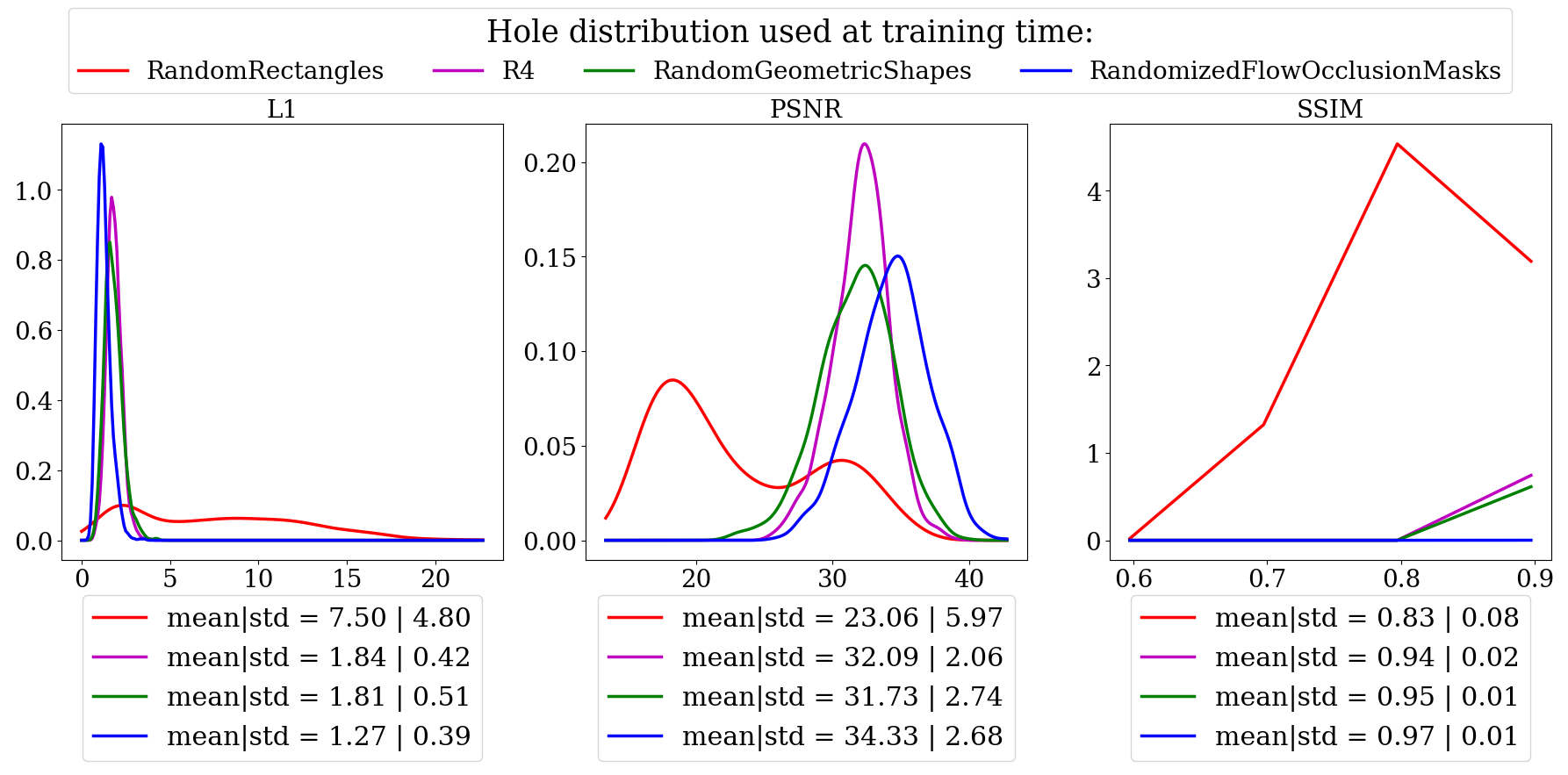}
        \label{F3_v_F20_r4}
    }
    \\
    \subfigure[Test w/ Random Geometric Shapes - $RGS$ hole configuration. $L_1$, $PSNR$, $SSIM$ histograms. Celeb-A. Left: 2+1 epochs. Right: 10+10 epochs.] 
    {
        \includegraphics[width=0.5\linewidth]{RGS_exp_3_sc_f32_faces_4way}
    	\rulesep
        \includegraphics[width=0.5\linewidth]{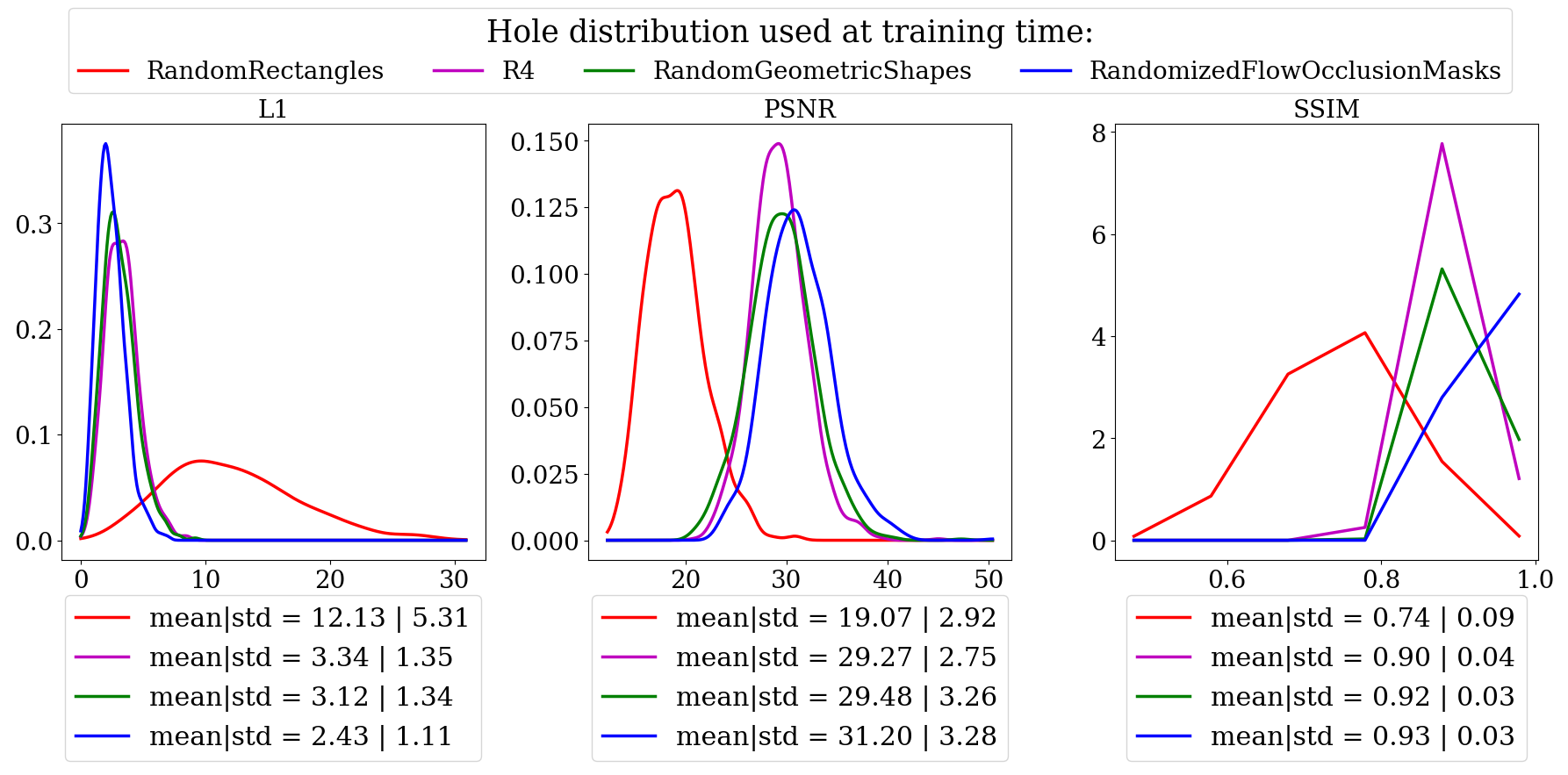}
        \label{F3_v_F20_rgs}
    }
    \\
    \subfigure[Test w/ Random Flow-based Occlusion/de-Occlusion Masks - $RFOM$ hole configuration. $L_1$, $PSNR$, $SSIM$ histograms. Celeb-A. Left: 2+1 epochs. Right: 10+10 epochs.] 
    {
        \includegraphics[width=0.5\linewidth]{RFOM_exp_3_sc_f32_faces_4way}
    	\rulesep
        \includegraphics[width=0.5\linewidth]{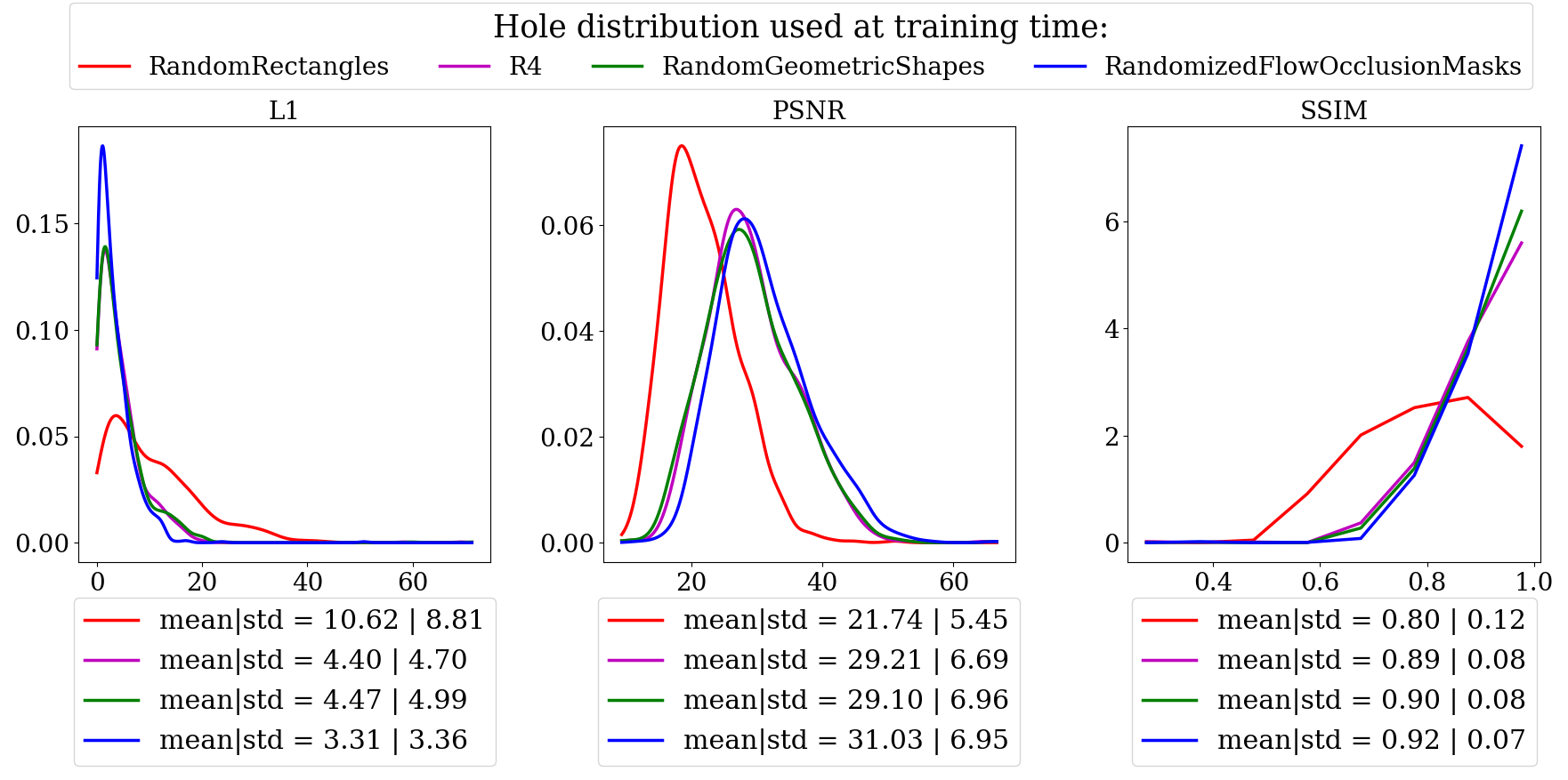}
        \label{F3_v_F20_rfom}
    }
    \caption{Smoothed empirical histograms of quantitative metrics evaluated in our train-test grid, for a 10.3MB in-painting neural network trained on on CUHK CelebA 90/10 split dataset. The left column figures are when the network is trained in 2+1 epochs (2 epochs of constant training rate and 1 of decaying rate). The right column figures are when network is trained in 10+10 epochs.  \textbf{Colors:} \textit{Red} represents training with $RR$, \textit{magenta} with $R4$, \textit{green} with $RGS$, and \textit{blue} with $RFOM$.}
    \label{fig:F3_v_F20}
\end{figure*}

Figure \ref{fig:F3_v_F20} gives cross-comparison quality metric histogram results when our network is trained on Celeb-A dataset. The left panel gives the results for when the network is trained in two epochs of constant training rate followed by one epoch of decaying training rate, i.e., a "$2+1$" training regime, while the right panel gives the results for a $10+10$ training epochs. These results support the hypothesis delineated in the paper. 

Notable observations are these: 
\begin{inparaenum}[(1)]
\item While hypothesis \ref{hyp:self} still holds, i.e. while $RR_{train}$ still has the best relative robustness results when tested  $RR$, the advantage of $RFOM$ becomes more clear with longer training. 
\item Similarly, with longer training $RGS$ develops greater robustness.
\item Longer training still clearly preserves the ability of Celeb-A to provide better robustness distinctions, i.e. hypothesis \ref{hyp:datasets} is still supported. Furthermore, if robustness comparison in short training runs are used to search the probability space of hole configurations, hypothesis \ref{hyp:datasets} remains useful.
\item Observing the position of $RR_{train}$ in all cross-comparison histogram indicates that hypothesis \ref{hyp:kernel}, regarding the impact of kernel-hole edge alignment in degrading robustness, still holds throughout.
\item Finally, observed relative robustness of $RGS_{train}$, with respect to variations in hole PDFs, seems most sensitive to the length of training. Networks $R4_{train}$ and $RFOM_{train}$ continue to have similar results but the superiority of $RFOM_{train}$ shows itself more clearly with longer training.  
\end{inparaenum}

The findings here mean that while shorter training runs might be useful in weeding out some lower-grade hole PDFs, longer training is required to distinguish the most robust hole configurations. This implies that complex hyper-parameter search methods should have varying regimes (length) of training at different stages in the search.  

\subsection{Changing Objectives on CelebA}


\begin{figure*}[h]
    \centering
    \subfigure[Test w/ RandomRectangles - $RR$ hole configuration. $L_1$, $PSNR$, $SSIM$ histograms. Left: Places ($L_1+PS+TV$ losses). Right: Celeb-A ($L_1+Dis$ losses).] 
    {
        \includegraphics[width=0.5\linewidth]{RR_exp_3_sc_f32_places_4way}
    	\rulesep
        \includegraphics[width=0.5\linewidth]{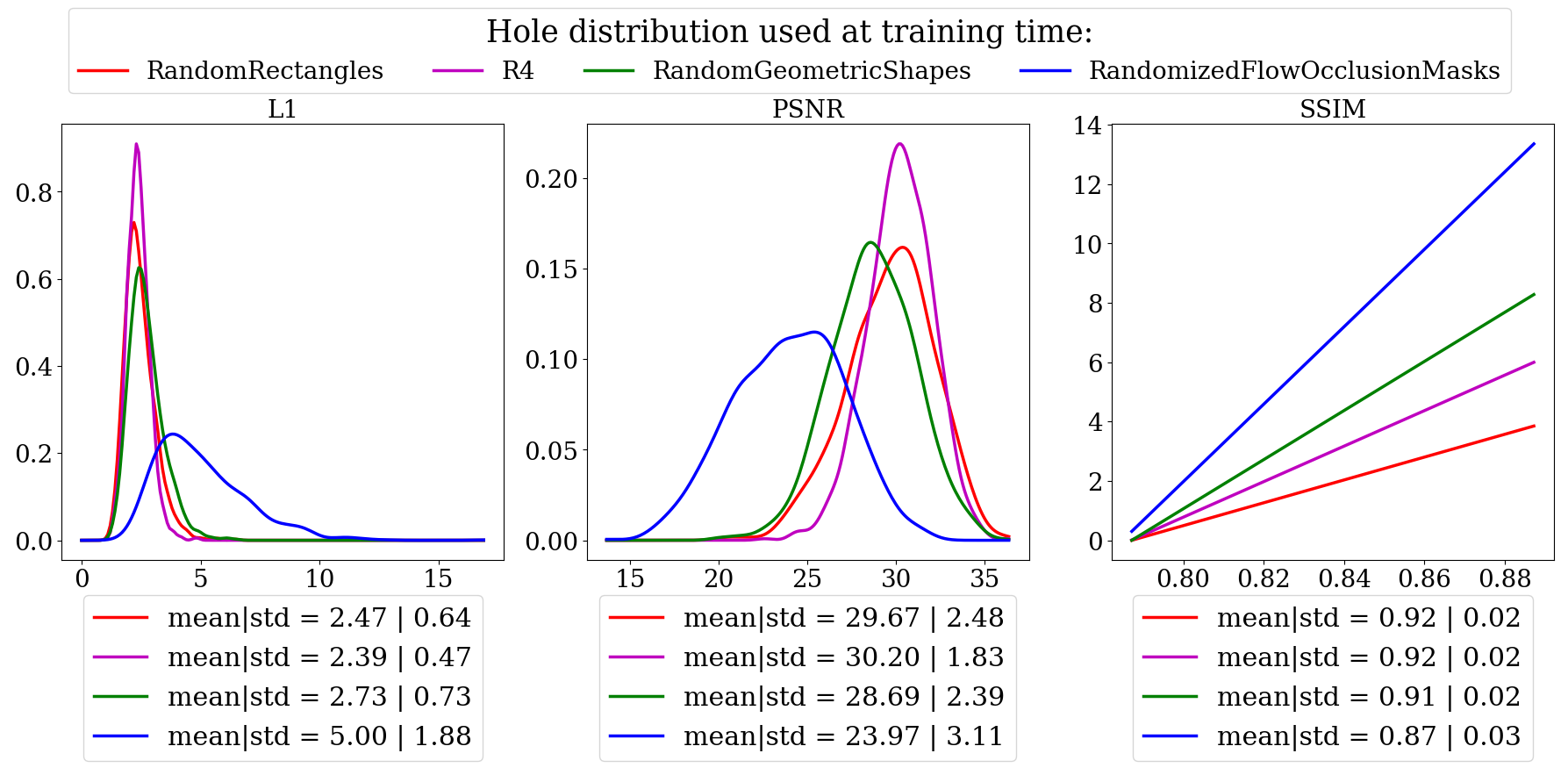}
        \label{p_f32_vs_f_f32_with_noPS_dis_rr}
    }
    \\
    \subfigure[Test w/ Randomly Rotated $RR$ - $R4$ hole configuration. $L_1$, $PSNR$, $SSIM$ histograms. Left: Places ($L_1+PS+TV$ losses). Right: Celeb-A ($L_1+Dis$ losses).] 
    {
        \includegraphics[width=0.5\linewidth]{R4_exp_3_sc_f32_places_4way}
    	\rulesep
        \includegraphics[width=0.5\linewidth]{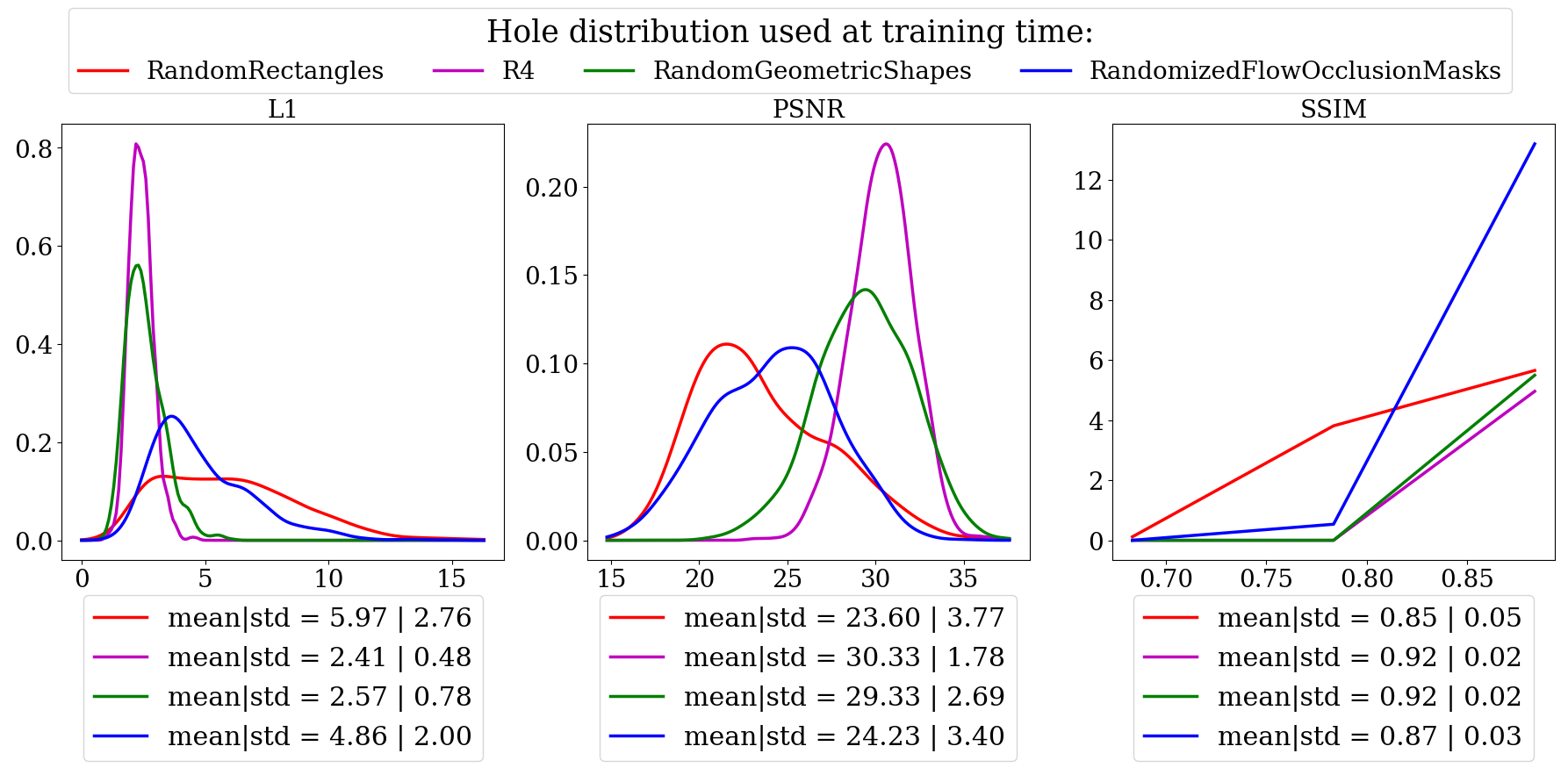}
        \label{p_f32_vs_f_f32_with_noPS_dis_r4}
    }
    \\
    \subfigure[Test w/ Random Geometric Shapes - $RGS$ hole configuration. $L_1$, $PSNR$, $SSIM$ histograms. Left: Places ($L_1+PS+TV$ losses). Right: Celeb-A ($L_1+Dis$ losses).] 
    {
        \includegraphics[width=0.5\linewidth]{RGS_exp_3_sc_f32_places_4way}
    	\rulesep
        \includegraphics[width=0.5\linewidth]{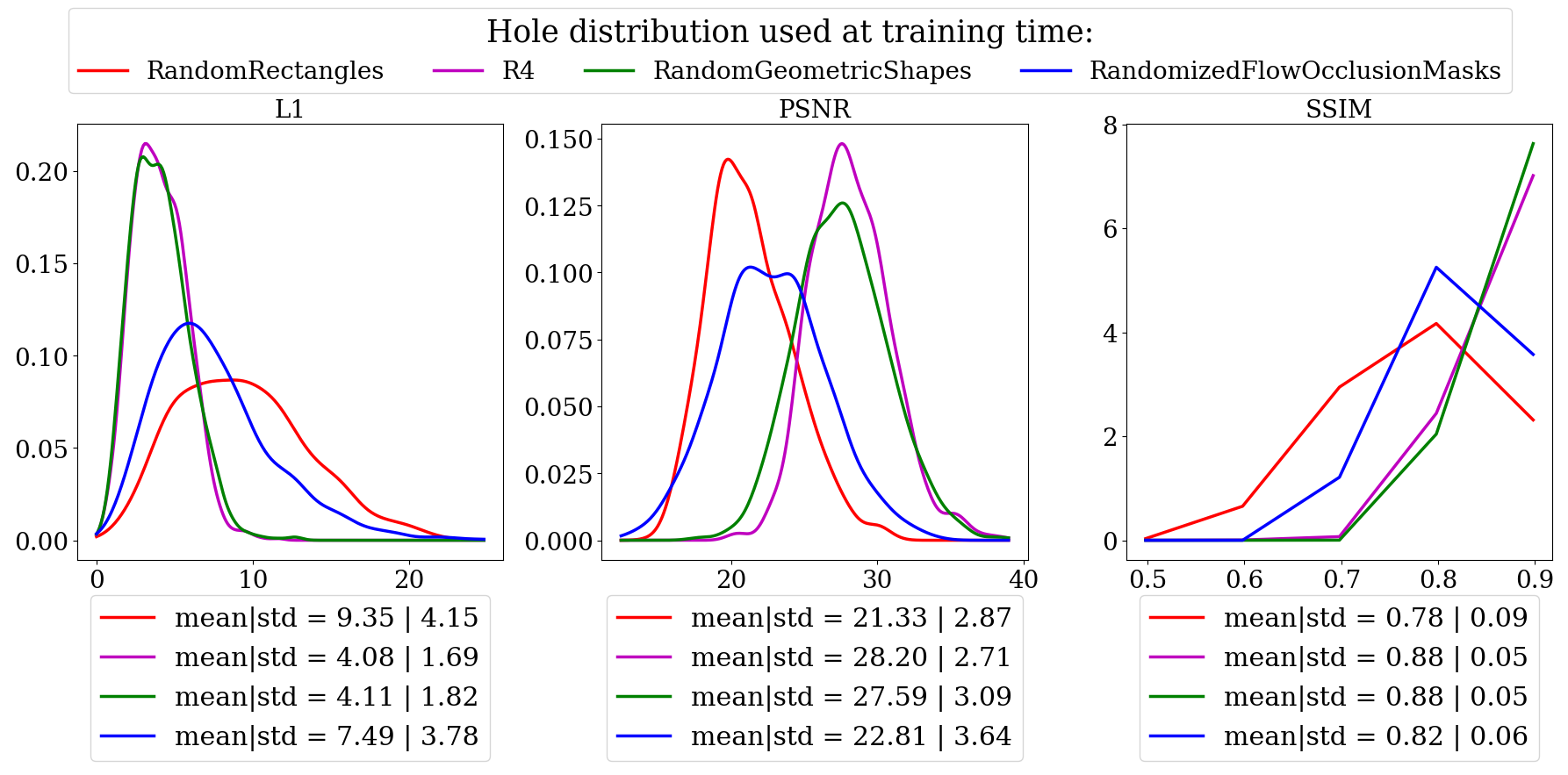}
        \label{p_f32_vs_f_f32_with_noPS_dis_rgs}
    }
    \\
    \subfigure[Test w/ Random Flow-based Occlusion/de-Occlusion Masks - $RFOM$ hole configuration. $L_1$, $PSNR$, $SSIM$ histograms. Left: Places ($L_1+PS+TV$ losses). Right: Celeb-A ($L_1+Dis$ losses).] 
    {
        \includegraphics[width=0.5\linewidth]{RFOM_exp_3_sc_f32_places_4way}
    	\rulesep
        \includegraphics[width=0.5\linewidth]{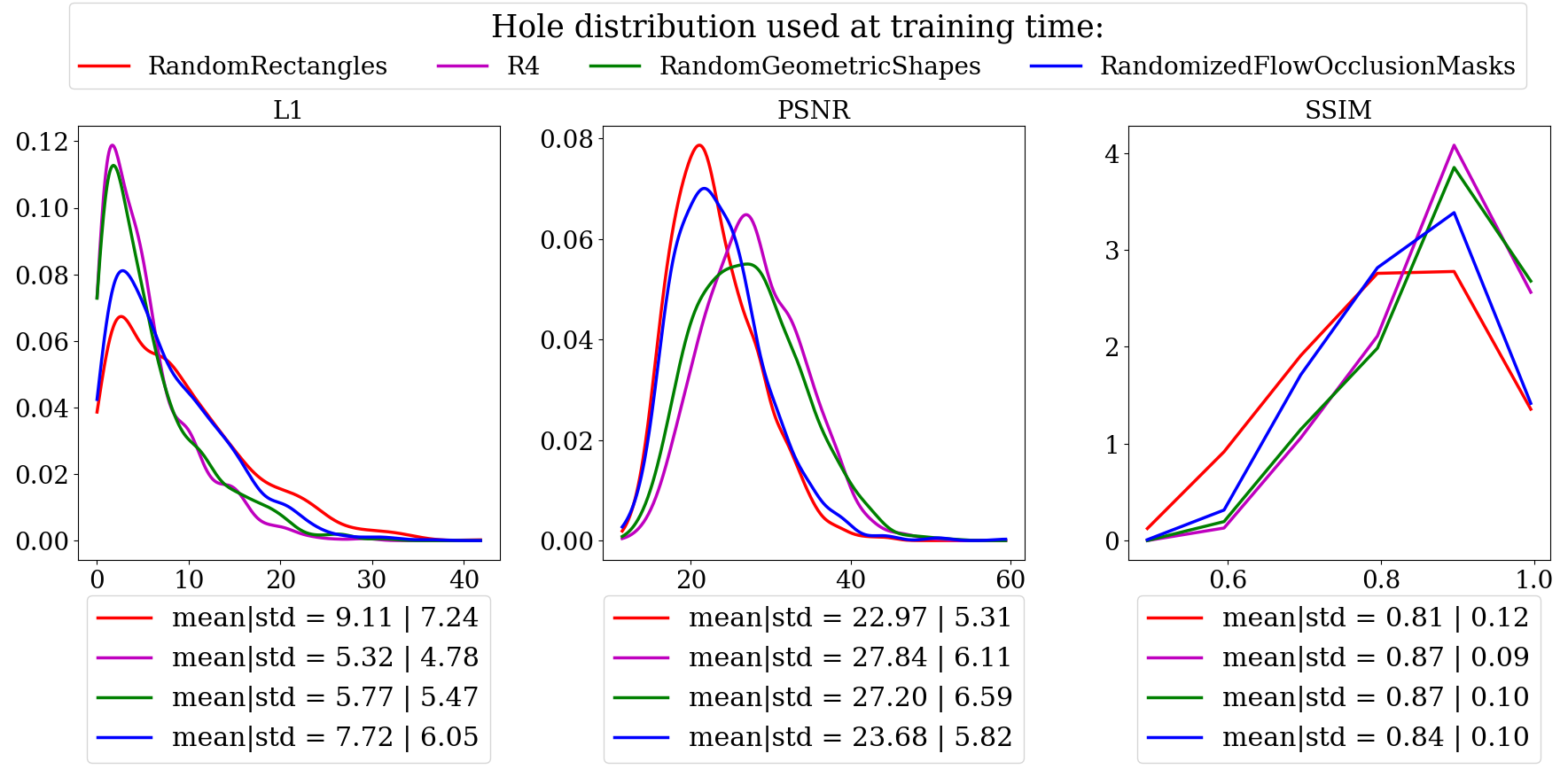}
        \label{p_f32_vs_f_f32_with_noPS_dis_rfom}
    }
    \caption{Smoothed empirical histograms of quantitative metrics evaluated in our train-test grid, for a 10.3MB in-painting neural network trained on MIT Places validation dataset (left column figures) and on CUHK CelebA \cite{liu2015faceattributes} full dataset (right column figures) and tested with each of the four hole configurations. The networks trained on Celeb-A, here, are trained with only $L_1$ and discriminative (GAN) losses. Training occurred for 2+1 epochs---at constant and at exponentially decaying learning rate. \textbf{Colors:} \textit{Red} represents training with $RR$, \textit{magenta} with $R4$, \textit{green} with $RGS$, and \textit{blue} with $RFOM$.}
    \label{fig:p_f32_vs_f_f32_with_noPS_dis}
\end{figure*}


\begin{figure*}[h]
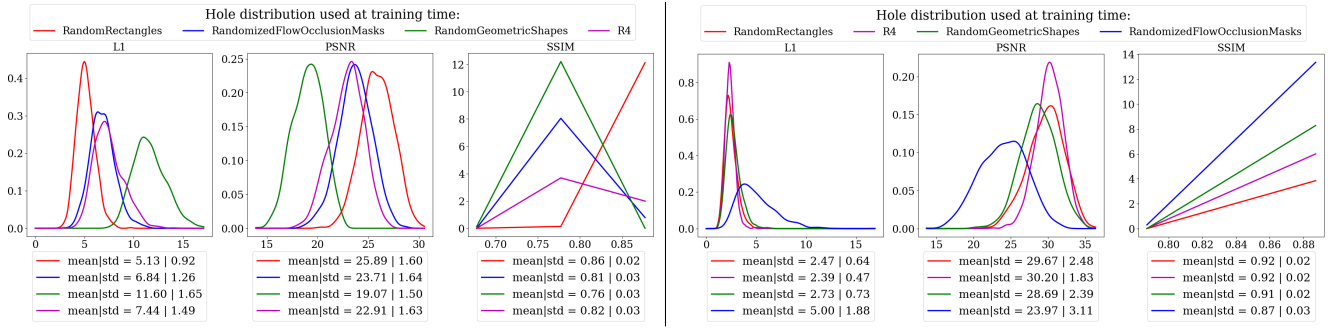
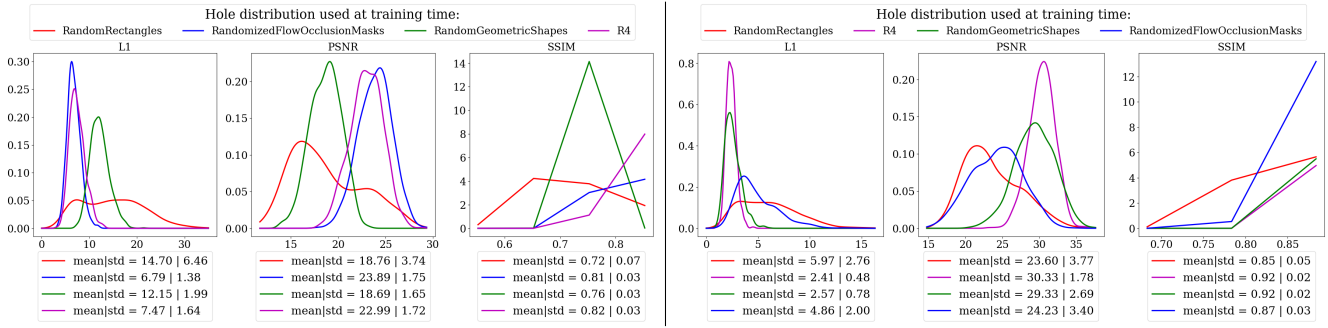
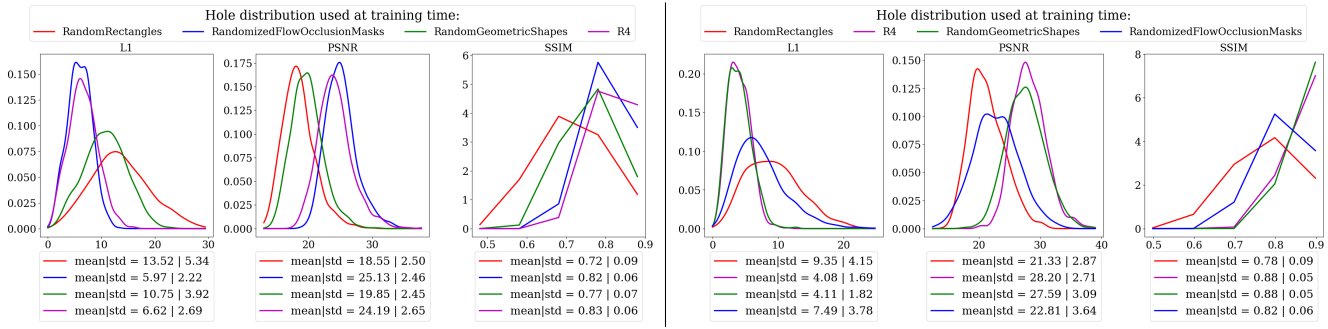
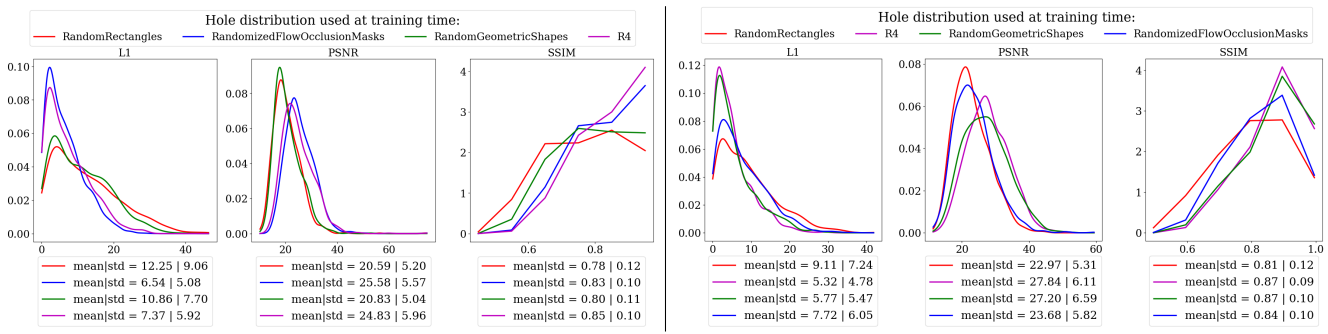

    \centering
    \subfigure[Test w/ RandomRectangles - $RR$ hole configuration. $L_1$, $PSNR$, $SSIM$ histograms. Celeb-A. Left: $L_1+PS+TV$. Right: $L_1+Dis$.] 
    {
        \includegraphics[width=0.5\linewidth]{RR_exp_3_sc_f32_faces_4way}
    	\rulesep
        \includegraphics[width=0.5\linewidth]{supp/RR_e3facesI2D1f32disnoPS}
        \label{e3facesI2D1f32disnoPS_rr}
    }
    \\
    \subfigure[Test w/ Randomly Rotated $RR$ - $R4$ hole configuration. $L_1$, $PSNR$, $SSIM$ histograms. Celeb-A. Left: $L_1+PS+TV$. Right: $L_1+Dis$.] 
    {
        \includegraphics[width=0.5\linewidth]{R4_exp_3_sc_f32_faces_4way}
    	\rulesep
        \includegraphics[width=0.5\linewidth]{supp/R4_e3facesI2D1f32disnoPS}
        \label{e3facesI2D1f32disnoPS_r4}
    }
    \\
    \subfigure[Test w/ Random Geometric Shapes - $RGS$ hole configuration. $L_1$, $PSNR$, $SSIM$ histograms. Celeb-A. Left: $L_1+PS+TV$. Right: $L_1+Dis$.] 
    {
        \includegraphics[width=0.5\linewidth]{RGS_exp_3_sc_f32_faces_4way}
    	\rulesep
        \includegraphics[width=0.5\linewidth]{supp/RGS_e3facesI2D1f32disnoPS}
        \label{e3facesI2D1f32disnoPS_rgs}
    }
    \\
    \subfigure[Test w/ Random Flow-based Occlusion/de-Occlusion Masks - $RFOM$ hole configuration. $L_1$, $PSNR$, $SSIM$ histograms. Celeb-A. Left: $L_1+PS+TV$. Right: $L_1+Dis$.] 
    {
        \includegraphics[width=0.5\linewidth]{RFOM_exp_3_sc_f32_faces_4way}
    	\rulesep
        \includegraphics[width=0.5\linewidth]{supp/RFOM_e3facesI2D1f32disnoPS}
        \label{e3facesI2D1f32disnoPS_rfom}
    }
    \caption{Smoothed empirical histograms of quantitative metrics evaluated in our train-test grid, for a 10.3MB in-painting neural network trained on on CUHK CelebA 90/10 split dataset. The left column figures are when the network is trained with $L1+PS+TV$ losses as given in \cite{liu2018partialinpainting}. The right column figures are when network is trained with discriminative losses.  Training occurred for 2+1 epochs---at constant and at exponentially decaying learning rate. \textbf{Colors:} \textit{Red} represents training with $RR$, \textit{magenta} with $R4$, \textit{green} with $RGS$, and \textit{blue} with $RFOM$.}
    \label{fig:faces_f32_PS_vs_dis}
\end{figure*}

We also tested the greater ability of the CUHK CelebA dataset to act as a relative robustness ``separator" by other means. (See section \ref{results}, Hypothsis H\ref{hyp:datasets} and H\ref{hyp:dataset_edges}.)

For example, we asked whether there was anything special about the optimization goals used? Would other optimization goals that would be more suited for the CelebA dataset produce different results?

In general, our in-painting experiments have shown us that use of (GAN) discriminative losses improve ability of the network to perform copies of patches, which seem to be more important in CelebA inpainting, given the general symmetry of faces. 
In fact, with Celeb-A inpainting, our inpainting network does better when discriminative losses are used instead of the ones used in general for all the results presented in the paper, i.e. \cite{liu2018partialinpainting}.

As such, it was important to test some of the hypothesis with a change in function estimation objectives used to train our in-painting network.

First, Figure \ref{fig:p_f32_vs_f_f32_with_noPS_dis} shows that even if we change the objective function for training on CelebA dataset it continues to remain a better ``separator" of relative robustness than the Places dataset (see hypothesis H\ref{hyp:datasets}), also under differing objectives used for CelebA inpainting.

We are only placing the results for Places dataset on the left-side panel of this figure to emphasize the contrast that shows what we mean by the ``separation" ability of the CelebA when it comes to relative robustness variation (expressed as shifts and ``separation" of the histograms) as we vary the holes used for training. 

The overlap of histograms continues to be far lower for CelebA. 

No further, specific comparison can be made, on the basis of this figure alone, between its right and left panels because not only the datasets but also the training objectives are entirely different. Again, the only reason they are placed next to each other is to highlight how the greater separation in histograms look in the right panel (CelebA) as opposed to the left panel (Places). 

So, we move to discussing the next, more important Figure \ref{fig:faces_f32_PS_vs_dis}. 

First, the expected bias towards holes used in training (hypothesis H\ref{hyp:self}) still holds.  

Whether we change the objective function from $L_1+PS+TV$ (as used \cite{liu2018partialinpainting}) or use some other objective function (e.g., here, GAN discriminative losses, $L_1+Dis$), CelebA continues to be a better separator of relative robustness, i.e., there is something inherent in the dataset (see Hypothesis H\ref{hyp:datasets}), and we have explored what this may be, in the section \ref{results} of the paper, through Hypothesis H\ref{hyp:dataset_edges}.

Given the results obtained with the differing objectives for CelebA (again, see Figure \ref{fig:faces_f32_PS_vs_dis}), one may formulate a new hypothesis which states the dependence of robustness on the in-painting objective function. We did not state this new hypothesis separately in the paper, because we have not yet explored this topic exhaustively enough. However, we provide some hints here because we believe it is an important refinement.  

While as we change training objectives for a network a dataset (e.g., CelebA) may retain its ``separation" ability (of in-painting relative robustness for various hole configurations), the (dominance) order of relative robustness may still be dependent on the specific objectives used.  

We also note that while the above paragraph may be stated as a new hypothesis, the behavior that expresses (and supports) it remains consistent with H\ref{hyp:self} to H\ref{hyp:datasets}.

To repeat the point, concretely, for purposes of clarity: 

There may be a reversal in the dominance order of relative robustness of various versions of a network (trained with a variety of hole configurations) when the network's training objectives change---as is evidenced by poorer performance of $RFOM_{train}$ observed in the comparison displayed in the left vs. the right panels of Figure \ref{fig:faces_f32_PS_vs_dis}. 

The conditions under which this reversal would dissipate or settle would be a subject worthy of further investigation.

\end{document}